\documentclass[letterpaper,twocolumn,10pt]{article}
\usepackage{usenix2019_v3}
\usepackage{xurl}
\usepackage{amsmath,amssymb,amsthm}
\usepackage{graphicx}
\usepackage{subfigure}
\usepackage{booktabs,array,multirow}
\usepackage{enumitem}
\usepackage{algorithm}
\usepackage[noend]{algpseudocode}
\usepackage{tcolorbox}
\tcbuselibrary{breakable,skins}
\algrenewcommand{\algorithmicforall}{\textbf{foreach}}
\algrenewcommand{\algorithmicindent}{1em}
\newtcolorbox[use counter=algorithm]{breakablealgorithm}[2]{
  enhanced jigsaw,breakable,sharp corners,
  colback=white,colframe=black,colbacktitle=white,coltitle=black,
  boxrule=0pt,toprule=0.5pt,bottomrule=0.5pt,titlerule=0.4pt,
  toprule at break=0.5pt,bottomrule at break=0.5pt,
  boxsep=0pt,left=6pt,right=6pt,top=2pt,bottom=3pt,
  lefttitle=4pt,righttitle=4pt,toptitle=2pt,bottomtitle=2pt,
  before skip=6pt,after skip=8pt,lines before break=5,
  fonttitle=\normalfont\normalsize,fontupper=\footnotesize,
  title={\textbf{Algorithm~\thealgorithm:} #1},label={#2},
  title after break={\textbf{Algorithm~\thealgorithm\ (continued)}}
}
\usepackage{float}
\usepackage{placeins}
\usepackage{wrapfig}
\usepackage{tikz}
\usetikzlibrary{arrows.meta}

\let\citep\cite
\makeatletter
\def\@lbibitem[#1]#2{\@bibitem{#2}}
\makeatother
\providecommand{\Description}[1]{}
\hypersetup{breaklinks=true,hypertexnames=false,hidelinks,
  pdfauthor={Xingyu Qu, Tianhao Lin, Yiqi Li, Zhiyu Chen, Sheng Wang},
  pdfsubject={arXiv preprint},pdflang={en}}
\date{}

\title{HAServe: Horizon-Aware Scheduling and Memory Co-Design for Efficient LLM Serving}
\author{
{\rm Xingyu Qu\textsuperscript{1}\quad Tianhao Lin\textsuperscript{1}\quad Yiqi Li\textsuperscript{1}\quad Zhiyu Chen\textsuperscript{2}\quad Sheng Wang\textsuperscript{1}}\\[4pt]
\textsuperscript{1}Wuhan University\\
\textsuperscript{2}Amazon
}
\hypersetup{pdftitle={HAServe: Horizon-Aware Scheduling and Memory Co-Design for Efficient LLM Serving}}
\begin{document}
\maketitle
\begingroup
\renewcommand{\thefootnote}{}
\footnotetext[0]{This manuscript updates the earlier PRISM version of this preprint. The revised system is named HAServe. Contact: Xingyu Qu \textless\href{mailto:whu-quxy@whu.edu.cn}{\texttt{whu-quxy@whu.edu.cn}}\textgreater}
\endgroup
\begin{abstract}
Online large language model (LLM) services process requests that share context but differ in prefix structure and arrival timing. Finite capacity couples request admission and key-value (KV) retention. This paper addresses one online serving problem: using a unified view of visible demand to choose the feasible joint admission and KV-retention candidate with the lowest predicted latency cost as prefix structures and arrival rates vary. To model this problem, we introduce HorizonSchema, a workload abstraction that connects exact prefix paths, permitted serializations, and released consumers as shared inputs to both decisions. Our analysis shows that useful reuse depends on demand and prefix survival, while service order changes latency even at equal computation. Furthermore, we present HAServe, which turns this abstraction into a scheduling--memory co-design method of systematic optimization for online LLM serving. We implement HAServe on SGLang and evaluate Qwen3-4B-Instruct-2507 and Llama2-13B on five workloads using eight H100 GPUs. Across the evaluated workloads, HAServe achieves up to $1.14$--$5.90\times$ P99 TTFT speedups and up to $1.08$--$1.44\times$ the throughput, with KV-cache hit rate up to 20.59 percentage points higher than the strongest baseline for each metric and setting.
\end{abstract}

\section{Introduction}
\label{sec:intro}

Online LLM services must process application context while maintaining responsive inference. Repeated computation over shared context consumes resources that could serve other requests. This context spans retrieved passages~\citep{lewis20rag,asai24selfrag}, conversation histories~\citep{zhao24wildchat,chiang24chatbotarena}, shared documents~\citep{li24loogle}, and agent instructions and tool definitions~\citep{yang24sweagent,xu25agentcompany}. Exact-prefix caching reuses key-value (KV) states to avoid repeated prefill~\citep{shazeer19fastdecoding,kwon23vllm,zheng24sglang,ye24chunkattention}. Requests differ in shared-prefix length, remaining prefill work, and arrival timing. These differences change the balance between computation savings, occupied KV capacity, and waiting. Efficient serving must adapt to current request demand.

Existing systems support prefix-aware execution and cache management~\citep{kwon23vllm,zheng24sglang,jin2025ragcache,ouyang26unicache}. Scheduling policies control batch admission~\citep{yu22orca,agrawal24sarathi,dexter25klpm,bari25slai}. Other approaches expose reusable prompt modules~\citep{gim24promptcache}, application dependencies~\citep{lin24parrot}, context structure~\citep{contextpilot2026}, or upcoming agent steps~\citep{pan25kvflow}. Our focus is the online coupling between request admission and KV retention. Admission determines which consumers run next and how long others wait. Retention determines which cached prefix states remain available and how much capacity is left for admission.

\begin{figure}[t]
\centering
\resizebox{\linewidth}{!}{
\begingroup
\definecolor{irSky}{HTML}{DCECF3}
\definecolor{irPeach}{HTML}{F8D4C4}
\definecolor{irMint}{HTML}{D7EDDF}
\definecolor{irRed}{HTML}{FF5733}
\definecolor{irGreen}{HTML}{34AA4E}
\begin{tikzpicture}[x=1pt,y=1pt,font=\sffamily\fontsize{8.7}{9.5}\selectfont,
 lab/.style={inner sep=0pt,align=center},
 title/.style={lab,font=\sffamily\fontsize{8.8}{9.5}\selectfont\bfseries\itshape},
 card/.style={draw=black,line width=.8pt,rounded corners=3pt},
 flow/.style={-{Stealth[length=3pt,width=2.6pt]},draw=black,line width=.8pt}]
\path[use as bounding box] (0,5) rectangle (190,95);

\draw[card,fill=irSky] (3,80) rectangle (188,94);
\node[lab] at (95.5,87) {Pending $q_2$: $AB\Vert u_2$ \quad KV: $AB,C$};
\draw[line width=.8pt] (6,80)--(6,16);
\draw[flow] (6,57)--(12,57);
\draw[flow] (6,16)--(12,16);

\node[title] at (101,72) {History-based Eviction};
\draw[card,fill=irPeach] (13,49) rectangle (101,65);
\node[lab] at (57,57) {Reclaim $AB$};
\draw[card,fill=black!4] (126,49) rectangle (188,65);
\node[title] at (164,57) {Miss};
\draw[flow] (101,57)--(124,57);
\fill[irRed] (138,57) circle (4.5pt);
\draw[white,line width=1.2pt,line cap=round] (136,55)--(140,59);
\draw[white,line width=1.2pt,line cap=round] (136,59)--(140,55);

\draw[irRed,dash pattern=on 4pt off 3pt,line width=.8pt] (15,42)--(188,42);

\node[title] at (101,34) {Demand-aware Retention};
\draw[card,fill=irPeach] (13,7) rectangle (101,25);
\node[lab] at (57,16) {Keep $AB$, reclaim $C$};
\draw[card,fill=irMint!55] (126,7) rectangle (188,25);
\node[title] at (164,16) {Hit};
\draw[flow] (101,16)--(124,16);
\fill[irGreen] (138,16) circle (4.5pt);
\draw[white,line width=1.2pt,line cap=round,line join=round] (135.5,16)--(137.3,14.2)--(141,18.2);
\end{tikzpicture}
\endgroup}
\caption{\textbf{Visible demand links prefix reuse, KV retention, and admission capacity.} Retaining $AB$ avoids a miss for $q_2$ if reclaiming $C$ leaves sufficient capacity. Here $u_2$ is the request-specific suffix.}
\Description{A shared input shows pending request q2 with path AB followed by private suffix u2, and cached paths AB and C. Two rows compare retention choices. History-based Eviction reclaims AB and leads to a miss, marked by a red cross. Demand-aware Retention keeps AB and reclaims C, allowing a hit when capacity permits, marked by a green check.}
\label{fig:intro_retention_choices}
\end{figure}

Figure~\ref{fig:intro_retention_choices} illustrates this coupling with two requests $q_1,q_2$ sharing prefix $AB$. After $q_1$ completes, $q_2$ arrives while $AB$ and an unrelated path $C$ remain cached. Keeping $AB$ avoids a miss for $q_2$ if reclaiming $C$ makes admission feasible. Serving $q_2$ next also changes other requests' waiting. Joint decisions must therefore weigh saved work, occupied capacity, and waiting.

\textbf{Our key idea is to use exact prefix paths and their visible consumers as the common input to admission and retention.} We introduce \emph{HorizonSchema} to capture arrival timing, exact token paths, and application-permitted serializations. At each decision, the \emph{visible horizon} is a bounded, arrival-ordered set of pending requests. Within this horizon, each path is linked to the requests that can consume it. This demand exists whether or not the path is cached. Backend state separately supplies active references, cache residency, and resource budgets. These inputs let the system compare serving choices over the same request set.

Building on HorizonSchema, controlled experiments and finite-horizon analysis establish three design requirements. First, locality decisions need both prefix structure and arrival timing. Their benefits change with demand and resource pressure. Second, retention must account for the consumers selected for service. Reclamation can change their missing work and reverse admission rankings. Third, the objective must count waiting across the visible horizon. Service order changes time to first token (TTFT) even at equal computation. A serving interval contributes latency to all requests still awaiting their first token. Computation savings for the selected batch must therefore be weighed against the delay imposed on the remaining queue. Together, these findings motivate evaluating admission and retention against the same requests and the same latency objective.

Guided by these requirements, we develop \textbf{HAServe} (\underline{H}orizon-\underline{A}ware \underline{Serv}ing) to coordinate scheduling and KV retention (Figure~\ref{fig:haserve_overview}). Its \emph{Horizon Scheduler} tracks visible requests and organizes them into prefix groups. It builds candidate batches and considers alternative serializations when applications permit reordering. For demand-aware candidates, the \emph{Demand-Aware Radix Tree} (DART) uses the proposed consumers to simulate reclamation. It returns prefix matches and available capacity after reclamation. The scheduler compares feasible joint choices using estimated mean remaining TTFT across the horizon. HAServe commits the selected batch and retention settings together. Runtime events refresh demand and residency for the next decision. This feedback loop coordinates computation reuse, memory allocation, and service order as online demand changes.

\begin{figure*}[!t]
\centering
\resizebox{0.80\linewidth}{!}{\input{figures/haserve_overview}}
\caption{\textbf{HAServe couples its Horizon Scheduler and Demand-Aware Radix Tree (DART) through shared prefix demand.} HorizonSchema links paths to visible consumers. Candidate demand guides retention, while resulting prefix matches and capacity guide admission.}
\Description{A left-to-right pipeline connects RAG, chat, document, and agent applications to two equally prominent core regions, HorizonSchema and HAServe. HorizonSchema represents exact prefix structure with a shared path that branches, and arrival timing with a clock. An arrow labelled Paths plus consumers supplies the Horizon Scheduler and DART inside HAServe. The scheduler coordinates admission with the retention feedback supplied by DART. A neural processor represents LLM serving, with prefill followed by decode and a response bubble. A storage cylinder represents the backend KV cache. A dashed feedback path returns runtime residency and capacity to HAServe, separately from the workload attributes in HorizonSchema.}
\label{fig:haserve_overview}
\end{figure*}

Our contributions are:
\begin{itemize}[leftmargin=*,topsep=1pt,itemsep=1pt,parsep=0pt,partopsep=0pt]
  \item We introduce HorizonSchema to express online prefix demand through exact paths, visible consumers, and arrival timing. Controlled experiments and finite-horizon analysis connect this demand to realized reuse and waiting (Section~\ref{sec:horizon_schema}).
  \item We design HAServe to coordinate prefix organization, request admission, and KV retention through the Horizon Scheduler and DART. It compares feasible joint choices using prefix matches and capacity after reclamation, together with estimated waiting across the same horizon (Section~\ref{sec:design}).
  \item We implement HAServe on SGLang and evaluate two models across five workloads on eight H100 GPUs. Across the evaluated workloads, HAServe achieves up to $1.14$--$5.90\times$ P99 TTFT speedups and up to $1.08$--$1.44\times$ the throughput. KV-cache hit rate is up to 20.59 percentage points higher. Each comparison uses the strongest baseline for the same metric and setting (Section~\ref{sec:evaluation}).
\end{itemize}


\section{Background}
\label{sec:background}

\subsection{Prompt Structure and Exact-Prefix Reuse}
\label{sec:abstraction}

Many online LLM services contain a shared prefix, reusable segments, and request-specific text. We index requests by $i$ and denote the $i$-th request by $q_i$. One possible serialization is
\begin{equation}
\label{eq:generic_prompt}
z_i = \Bigl[x^{\mathrm{sys}} \,\Vert\, x^{r_{i,1}} \,\Vert\, x^{r_{i,2}} \,\Vert\, \cdots \,\Vert\, x^{r_{i,m_i}} \,\Vert\, x_i^{\mathrm{priv}}\Bigr],
\end{equation}
where $\Vert$ denotes concatenation, $x^{\mathrm{sys}}$ is a shared system prompt, $r_{i,j}$ identifies reusable segment $j$ in request $i$, $m_i$ is the number of reusable segments in $q_i$, and $x_i^{\mathrm{priv}}$ is the request-specific suffix. The segments may be retrieved passages, document blocks, or conversation history. Their order is part of the token sequence: only an application that permits a particular reordering can expose it as a serving action, and preserving segment content alone does not establish answer equivalence, because evidence position can affect model responses~\citep{liu24lostmiddle}.

\subsection{Serving Latency and Resource Constraints}
\label{sec:ttft_decomposition}

For request $q_i$, we decompose TTFT as
\begin{equation}
\label{eq:ttft_decomp}
\mathrm{TTFT}_i
=
 W_i^{\mathrm{admit}}
+ T_i^{\mathrm{prefill}}
+ T_i^{\mathrm{1tok}},
\end{equation}
where $W_i^{\mathrm{admit}}$ is the admission wait from the arrival of $q_i$ to the backend prefill launch, $T_i^{\mathrm{prefill}}$ is the prefill interval that builds missing KV states, and $T_i^{\mathrm{1tok}}$ is the time from prefill completion to the first returned token. Admission wait includes scheduler-side waiting, backend enqueue, and waiting for GPU service.

Let $\mathcal{C}_t$ denote the KV-cached prefix paths resident when $q_i$ is admitted at time $t$. Native exact-prefix caching matches tokens together with their preceding context~\citep{zheng24sglang}. Under a fixed model and tokenization, the reusable hit length is
\begin{equation}
\label{eq:hit_length}
L_i^{\mathrm{hit}}(t)=\max_{c \in \mathcal{C}_t} \operatorname{LCP}(z_i, c),
\end{equation}
where $\operatorname{LCP}$ counts the longest common prefix tokens and the maximum is zero for an empty cache. With $\ell(\cdot)$ denoting token length, the missing-token count is $\ell(z_i)-L_i^{\mathrm{hit}}(t)$, giving the prefill-time approximation
\begin{equation}
\label{eq:prefill_time}
T_i^{\mathrm{prefill}}
\approx
\frac{\ell(z_i)-L_i^{\mathrm{hit}}(t)}{R_{\mathrm{pf}}(B_t)},
\end{equation}
where $R_{\mathrm{pf}}(B_t)$ is the realized prefill throughput under batch state $B_t$. Segment identities are only control-plane hints; KV hits still require exact serialized-prefix equality. Reuse can lower the prefill term, but a policy that delays admission or changes batch composition can increase other TTFT terms. Scheduling and cache retention must therefore be assessed together rather than by hit rate alone.

\noindent\textbf{Compute and KV memory.} The data plane is constrained by both prefill compute and KV capacity~\citep{kwon23vllm,agrawal24sarathi}. The missing token length in Eq.~\eqref{eq:prefill_time} determines prefill work, while retaining longer or more numerous prefixes consumes KV blocks needed by other requests and active decode. A retained prefix must survive until a matching request uses it. Eviction during that request's wait can still create a miss. The relative costs vary with prompt length, batch state, available KV blocks, and the requests presently waiting.


\section{HorizonSchema: Workload Abstraction and Design Motivation}
\label{sec:horizon_schema}
\label{sec:bottleneck}

HorizonSchema links exact prefix paths to their visible consumers and arrival times. Three controlled experiments motivate using this representation to coordinate prefix organization, admission, and retention in Section~\ref{sec:design}.

\subsection{Request Structure and Arrival Timing}
\label{sec:request_forms}
\label{sec:horizon_load}

Table~\ref{tab:request_forms} summarizes four common request patterns. Their reusable paths differ in ordering constraints, KV footprint, and remaining prefill work; several patterns may coexist in one service.

\begin{table*}[!t]
\centering
\caption{\textbf{Request structure shapes reuse and resource demand.} Four representative patterns and their serving implications.}
\label{tab:request_forms}
\small
\setlength{\tabcolsep}{4pt}
\renewcommand{\arraystretch}{1.08}
\begin{tabular}{@{}>{\raggedright\arraybackslash}p{0.24\linewidth}>{\raggedright\arraybackslash}p{0.32\linewidth}>{\raggedright\arraybackslash}p{0.39\linewidth}@{}}
\toprule
Request pattern & Reusable structure & Serving implication \\
\midrule
Retrieved passages & Partially overlapping passages & Exact overlap depends on passage order and preceding tokens \\
Conversation history & An expanding session prefix & Reuse depends on released turns and the gap between them \\
Shared long document & A stable prefix across questions & Large prefill savings compete with a large KV footprint \\
Policy and tool history & Shared templates with task histories & Template hits can leave substantial task-specific work \\
\bottomrule
\end{tabular}
\end{table*}

\begin{figure*}[!t]
\centering
\resizebox{0.82\linewidth}{!}{
\definecolor{hsTeal}{HTML}{258875}
\definecolor{hsMint}{HTML}{70C9B6}
\definecolor{hsBlue}{HTML}{4595D2}
\definecolor{hsSky}{HTML}{78B9EE}
\definecolor{hsAmber}{HTML}{F6B84A}
\begin{tikzpicture}[x=1pt,y=1pt,font=\fontsize{8.7}{10}\selectfont,
 lab/.style={inner sep=0pt,align=center},
 title/.style={lab,font=\fontsize{9}{10.5}\selectfont\bfseries},
 block/.style={draw=black!55,line width=.6pt,minimum height=30pt,inner sep=2pt,align=center},
 flow/.style={-{Stealth[length=3.5pt,width=2.7pt]},draw=black!60,line width=.65pt},
 attach/.style={densely dotted,draw=black!45,line width=.6pt}]
\path[use as bounding box] (0,0) rectangle (392,106);
\fill[hsMint!10,rounded corners=2pt] (82,1) rectangle (268,105);
\fill[hsTeal!55] (82,103) rectangle (268,105);
\node[title] at (175,93) {HorizonSchema $\mathcal W=(\mathcal T,\mathcal P)$};
\node[block,fill=black!3,minimum width=62pt] (requests) at (31,65) {Released\\requests};
\node[block,fill=hsSky!35,minimum width=72pt] (timing) at (126,65) {Arrival timing\\$\mathcal T$};
\node[block,fill=hsMint!45,minimum width=72pt] (prefix) at (222,65) {Prefix structure\\$\mathcal P$};
\draw[flow] (requests.east)--(82,65);
\fill[hsSky!8,rounded corners=2pt] (289,24) rectangle (392,105);
\node[title] at (340,93) {Visible demand};
\node[block,fill=hsMint!25,minimum width=88pt,minimum height=20pt] (paths) at (340.5,74) {Prefix paths};
\node[block,fill=hsSky!20,minimum width=88pt,minimum height=20pt] (consumers) at (340.5,49) {Consumers};
\draw[draw=black!60,line width=.65pt] (268,65)--(281,65);
\draw[flow] (281,65)|-(paths.west);
\draw[flow] (281,65)|-(consumers.west);
\node[lab] at (340.5,30) {Within $\mathcal H_t$};
\draw[attach] (timing.south)--(126,41);
\node[lab] at (126,34) {Release time, age};
\draw[flow] (101,17)--(152,17);
\foreach \x in {108,119,139}{\fill[hsBlue] (\x,17) circle (1.8pt);}
\draw[attach] (prefix.south)--(222,41);
\node[lab] at (222,34) {Exact token paths};
\fill[hsAmber!25,rounded corners=1.5pt] (174,5) rectangle (263,24);
\node[lab,font=\fontsize{8.5}{10}\selectfont] at (218.5,14.5) {Legal orders $\mathcal L_i$};
\node[lab,font=\fontsize{8.5}{10}\selectfont] at (305,14.5) {Optional};
\draw[attach] (264,14.5)--(282,14.5);
\end{tikzpicture}}
\caption{\textbf{HorizonSchema links prefix paths to their visible consumers.} Arrival timing, exact token paths, and optional legal orders identify prefix paths and the requests that demand them within $\mathcal H_t$.}
\Description{A compact left-to-right abstraction flow connects released requests to HorizonSchema and then to visible demand. HorizonSchema has two parallel attribute modules: arrival timing T, with release time and age, and prefix structure P, with exact token paths. An amber legal-orders annotation is optional. A branching connector leads to two separate modules, Prefix paths and Consumers, grouped under Visible demand within the visible horizon. They represent exact prefix sequences and the requests whose considered serializations begin with those sequences, independently of cache residency. The diagram contains no admission, cache-reclamation, or serving execution stages.}
\label{fig:horizon_abstraction}
\end{figure*}

Arrival timing determines which matching consumers coexist and how long a cached prefix must survive before use. Production workloads vary in request intensity, inter-arrival times, and request lengths~\citep{wang25ragpulse,xiang26servegen,nixon26year}. Even at the same arrival rate, different request interleavings can change pending demand and cache trajectories. HorizonSchema therefore links each released request's arrival time to its exact token path.

\subsection{HorizonSchema and the Visible Horizon}
\label{sec:schema_components}
\label{sec:observable_horizon}

We write \emph{HorizonSchema} as $\mathcal W=(\mathcal T,\mathcal P)$ (Figure~\ref{fig:horizon_abstraction}). Both components use the same request identifiers: $\mathcal T$ records arrival timing, while $\mathcal P$ records exact token paths and application-permitted alternatives.

\noindent\textbf{Example.} Consider two pending RAG requests $q_1,q_2$ arriving at $t_1<t_2$, with token paths $z_i=[x^{\mathrm{sys}}\Vert D\Vert u_i]$. They share a system prompt and document $D$, followed by private queries $u_i$. The records link each arrival time to its token path. At $t\ge t_2$, their ages are $t-t_1$ and $t-t_2$, and both are consumers of the shared prefix (Figure~\ref{fig:horizon_example}).

\begin{figure}[!htbp]
\centering
\resizebox{\linewidth}{!}{
\begingroup
\definecolor{heBadge}{HTML}{5345FF}
\definecolor{hePeach}{HTML}{F7D5B8}
\definecolor{heMint}{HTML}{C8E5DD}
\definecolor{heSky}{HTML}{ADD3E3}
\begin{tikzpicture}[x=1pt,y=1pt,font=\sffamily\fontsize{8.3}{9.5}\selectfont,
 lab/.style={inner sep=0pt,align=center},
 title/.style={lab,font=\sffamily\fontsize{8.6}{10}\selectfont\bfseries\itshape},
 card/.style={draw=black,line width=.75pt,rounded corners=3pt},
 token/.style={draw=black!80,line width=.5pt,rounded corners=1.4pt,minimum height=15pt,inner sep=1pt,align=center},
 badge/.style={circle,fill=heBadge,text=white,inner sep=0pt,minimum size=12pt,font=\sffamily\fontsize{8}{9.5}\selectfont\bfseries},
 flow/.style={-{Stealth[length=3.5pt,width=2.8pt]},draw=black,line width=.75pt}]
\path[use as bounding box] (-6,-3) rectangle (236,141);
\draw[card,fill=black!5] (0,65) rectangle (81,132);
\node[title] at (40.5,117) {Arrival timing $\mathcal T$};
\draw[flow] (9,94)--(73,94);
\fill[black] (20,94) circle (1.6pt);
\fill[black] (43,94) circle (1.6pt);
\draw[densely dashed,line width=.55pt] (65,88)--(65,106);
\node[lab] at (20,104) {$q_1$};
\node[lab] at (43,104) {$q_2$};
\node[lab] at (20,81) {$t_1$};
\node[lab] at (43,81) {$t_2$};
\node[lab] at (65,81) {$t$};
\node[badge] at (0,132) {1};
\draw[card,fill=hePeach] (91,65) rectangle (236,132);
\node[title] at (163.5,117) {Prefix structure $\mathcal P$};
\node[lab] at (100,96) {$q_1$};
\node[lab] at (100,76) {$q_2$};
\node[token,fill=heSky,minimum width=20pt] at (119,96) {$x^{\mathrm{sys}}$};
\node[token,fill=heMint,minimum width=18pt] at (141,96) {$D$};
\node[token,fill=white,minimum width=76pt] at (193,96) {What is X?};
\node[token,fill=heSky,minimum width=20pt] at (119,76) {$x^{\mathrm{sys}}$};
\node[token,fill=heMint,minimum width=18pt] at (141,76) {$D$};
\node[token,fill=white,minimum width=76pt] at (193,76) {Explain Y.};
\node[badge] at (91,132) {2};
\draw[line width=.75pt] (40.5,65)--(40.5,58)--(163.5,58)--(163.5,65);
\draw[flow] (118,58)--(118,51);
\draw[card,fill=heMint] (0,0) rectangle (236,51);
\node[title] at (118,39) {Visible consumers};
\node[lab] at (118,24) {$[x^{\mathrm{sys}}\Vert D]\ \longrightarrow\ \{q_1,q_2\},\qquad q_1,q_2\in\mathcal H_t$};
\node[badge] at (0,51) {3};
\node[lab,font=\sffamily\fontsize{8.1}{9.5}\selectfont] at (118,9) {Cached or not: the same demand};
\end{tikzpicture}
\endgroup}
\caption{\textbf{One prefix, two visible consumers.} Both requests are in $\mathcal H_t$ at $t\ge t_2$. Their shared system prompt $x^{\mathrm{sys}}$ and document $D$ define demand independently of cache residency.}
\Description{A compact two-row diagram uses three pastel rounded boxes, black arrows, and blue numbered badges. Box 1 shows q1 arriving at t1 and q2 at t2 before decision time t. Box 2 shows their complete token paths, with identical system-prompt and document blocks followed by What is X? and Explain Y. Box 3 links the exact system-prompt/document prefix to consumers q1 and q2, explicitly within the visible horizon. A note states that cached or uncached prefixes have the same demand.}
\label{fig:horizon_example}
\end{figure}

\noindent\textbf{Arrival timing.} $\mathcal T$ records released request identifiers and timestamps, providing waiting ages, inter-arrival gaps, and the recent arrival rate $\widehat\lambda_t$. Unreleased requests contribute no visible demand.

\noindent\textbf{Prefix structure.} $\mathcal P$ records each released request's token path $z_i$. Optional annotations include a segment skeleton $\Pi_i$, a set $\mathcal L_i$ of application-permitted complete serializations, and a predecessor link. Without permitted alternatives, $\mathcal L_i=\{z_i\}$. HAServe selects among these legal paths when constructing candidates.

Permitted reordering can change exact prefix matches even when segment contents are unchanged. Its answer-quality effects are evaluated separately.

\noindent\textbf{Visible demand.} Let $\mathcal Q_t$ denote the pending requests at time $t$. The \emph{visible horizon} $\mathcal H_t\subseteq\mathcal Q_t$ is a bounded, arrival-ordered subset, fixed before comparing serving choices from the same runtime snapshot. It adds no collection delay to await future requests.

A \textbf{prefix path} is the exact token sequence from the prompt's start to a chosen prefix boundary. Its \textbf{visible consumers} are requests in $\mathcal H_t$ whose considered serialization begins with that sequence. This relation expresses prefix demand independently of cache residency and does not guarantee a hit. A known predecessor link does not expose an unreleased successor as a consumer. The scheduler determines which consumers run next, while cache retention determines which resident prefix paths remain available to them.

\noindent\textbf{Decision state.} Fix $\mathcal H_t$, its request records, and runtime snapshot $s_t=(\mathcal Q_t,\mathcal A_t,\mathcal C_t,b_t)$, where $\mathcal A_t$ contains active requests, $\mathcal C_t$ resident KV paths, and $b_t$ compute and KV budgets. A candidate $a$ chooses the next batch, legal paths $z_i^a\in\mathcal L_i$, and retention under backend constraints. Candidates use a common FIFO/LRU continuation for the remaining horizon. Section~\ref{sec:joint_selection} ranks them by predicted mean remaining TTFT; only the current action is committed.

\subsection{Observations and Design Implications}
\label{sec:motivation_observations}
\label{sec:queue_weighted}
\label{sec:horizon_cost}
\label{sec:visible_work}

Three experiments examine demand, residency, and waiting. The first tests when locality-oriented admission helps. The second connects aggregate path residency to an event in which a waiting consumer loses its cached prefix. The third isolates the latency cost of service order at equal computation.
All three experiments use Qwen3-4B. First-come, first-served (FCFS) provides the admission reference. The longest-prefix matching (LPM) scheduling policy prioritizes requests with longer currently resident prefix matches. Least-recently-used (LRU) and least-frequently-used (LFU) provide retention rules. Each paired comparison uses identical request contents, arrival timestamps, and resource budgets. We follow the five-seed measurement protocol in Section~\ref{sec:eval:setup}. Cache states then evolve under the selected policies.

\begin{figure}[!htbp]
\centering
\includegraphics[width=\linewidth]{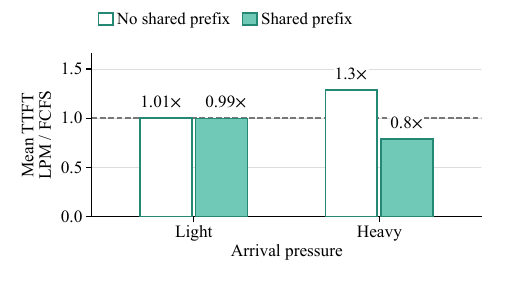}
\caption{\textbf{Locality benefits depend on prefix sharing and arrival pressure.} Bars show LPM/FCFS mean TTFT under light and heavy arrivals, with and without a shared prefix. The dashed line marks equal mean TTFT.}
\Description{Grouped bars compare LPM-to-FCFS mean TTFT under light and heavy arrival pressure, with and without shared prefixes. The dashed line marks equal mean TTFT.}
\label{fig:motivation_demand}
\end{figure}

\noindent\textbf{Observation 1. The benefit of prefix locality depends on both request structure and arrival pressure.}
We use synthetic prompts with a mean length of 4,096 tokens and vary shared-prefix length between 0 and 2,048 tokens under LRU retention. Light and heavy arrivals use $\lambda/\mu_{\mathrm{ref}}=0.30$ and $0.95$, where $\lambda$ is the offered rate and $\mu_{\mathrm{ref}}$ is an independently calibrated reference service rate. Each run has 1,000 warmup requests followed by 10,000 measured requests (Figure~\ref{fig:motivation_demand}).

Under light pressure, LPM has little effect on mean TTFT in either prefix condition. Under heavy pressure, it increases mean TTFT by roughly 30\% without a shared prefix but reduces it by roughly 20\% with one. In the latter setting, P99 TTFT nevertheless increases by 49.3\%. \textbf{Neither arrival pressure nor prefix length alone identifies a beneficial admission choice, and an average improvement need not extend to the tail.} Admission therefore needs the structure and timing of visible requests to assess useful reuse under current demand.

\noindent\textbf{Observation 2. Scheduling and KV retention jointly determine realized prefix reuse.}
To examine path availability, we combine FCFS and LPM with LRU and LFU on five document-disjoint LooGLE cohorts~\citep{li24loogle}. Each cohort contains eight documents, with 32 warmup requests followed by 64 measured requests. We hold each cohort's request trace fixed and vary the KV budget in 16-token blocks. Capacity pressure $\chi$ is the number of distinct reusable prefix blocks divided by KV capacity in blocks. We target $\chi\in\{0.5,1,2,4\}$, with larger $\chi$ indicating tighter capacity.

\begin{figure}[!htbp]
\centering
\includegraphics[width=\linewidth]{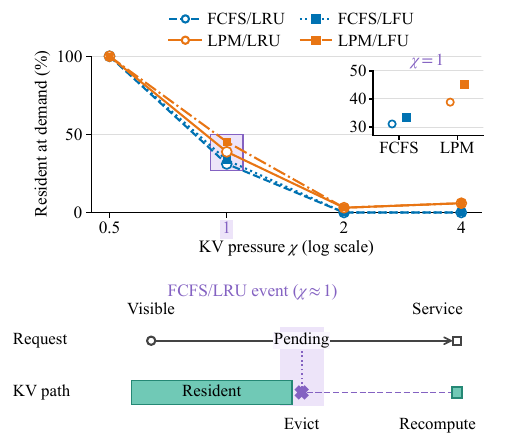}
\caption{\textbf{Scheduling and retention jointly determine prefix availability.} Top: residency at demand pooled across five cohorts; inset: $\chi=1$. Bottom: a measured FCFS/LRU event with a matching consumer pending at eviction. Positions show event order, not elapsed time.}
\Description{Four curves compare residency at demand under FCFS and LPM scheduling with LRU and LFU retention. An inset separates the four measured values at unit KV pressure. Below, separate request and KV-path lanes show a consumer becoming visible while its prefix is resident, eviction while the consumer is pending, and recomputation when it receives service.}
\label{fig:motivation_residency}
\end{figure}

Figure~\ref{fig:motivation_residency} reports residency at demand, the fraction of demands for previously inserted blocks that find those blocks still resident at cache match. With ample capacity, no measured eviction occurs and all four policies perform the same missing work within each cohort. As the reusable working set approaches capacity, LPM improves residency and reduces missing work under both retention rules, with a larger improvement under LFU. Under tighter capacity, most previously built paths are absent when demanded, and the benefits do not grow monotonically with pressure.

Event records expose the mechanism behind these aggregate changes. The event sequence in Figure~\ref{fig:motivation_residency} follows a previously built prefix block through physical eviction and recomputation, with no intervening reinsertion. Its matching consumer was already pending before eviction. The path was therefore reclaimed despite visible demand for it, identifying an opportunity to use current demand in retention decisions.

Scheduling changes when paths are built and consumed, while retention determines which of them survive those intervals. Retained paths also occupy space needed to admit other requests. At $\chi\approx1$, reductions in the interval from backend receipt to the first prefill launch account for about 84\% and 78\% of the aggregate mean TTFT reductions from FCFS to LPM under LRU and LFU, respectively. These percentages are ratios of mean differences across the five cohorts. The interval includes queueing and resource waits. \textbf{Scheduling and retention therefore need a shared view of consumers, consumption order, and available capacity to turn prefix locality into realized reuse.}

\begin{figure}[!htbp]
\centering
\includegraphics[width=\linewidth]{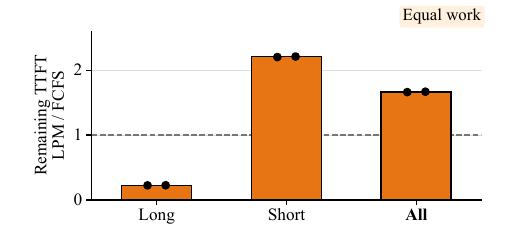}
\caption{\textbf{Service order changes latency even at equal computation.} Long requests with prefix hits are queued behind short requests. Bars average the paired LPM/FCFS remaining-TTFT ratios for long requests, short requests, and all requests; dots show individual runs.}
\Description{Three bars compare LPM-to-FCFS remaining TTFT for long requests with prefix hits, short requests, and all requests under equal computation. Each bar shows the mean paired ratio with individual-run dots. The dashed line marks equal remaining TTFT.}
\label{fig:motivation_waiting}
\end{figure}

\noindent\textbf{Observation 3. Prioritizing prefix hits can increase latency even when computation is unchanged.}
To separate waiting effects from recomputation, we queue 112 short requests ahead of 16 long hot requests and hold dispatch until all are pending. Each hot request contains 16,384 input tokens with an 8,192-token resident prefix, leaving 8,192 tokens to prefill. Each short request has 512 input tokens and no prefix hit. FCFS and LPM see the same per-request hits and missing work, with no measured eviction. Remaining TTFT is measured from the common release point to the first returned token (Figure~\ref{fig:motivation_waiting}).

{\widowpenalty=10000
LPM substantially reduces hot-request latency, but more than doubles short-request latency and raises the overall mean by about two-thirds. When hot requests are queued first, the overall difference is small. A large prefix hit therefore does not imply little remaining work. \textbf{Advancing a few long requests can delay many short ones without saving any computation.} Admission must therefore assess the waiting imposed on other visible requests together with the work it advances.\par}

\noindent\textbf{Connecting work to waiting.} A service interval delays every horizon request still awaiting its first token. Saved prefix work therefore matters through the waiting it removes, and candidate-specific reclamation can change both missing work and admission rankings. Appendix~\ref{sec:appendix:queueing} gives the latency accounting and a ranking-reversal example; Section~\ref{sec:joint_selection} uses this accounting to compare feasible candidates.

\noindent\textbf{Design implications.}
\label{sec:design_implications}
The observations establish three requirements for serving:
\begin{enumerate}[label=(\arabic*),leftmargin=*,topsep=2pt,itemsep=2pt]
  \item Represent current demand through both exact prefix structure and request timing.
  \item Tie retention priorities to the consumers selected for service.
  \item Evaluate serving choices with a latency objective that counts waiting across the visible horizon.
\end{enumerate}
HAServe uses \textbf{exact prefix paths and visible consumers as the common input to admission and retention}, evaluating both decisions against the same latency objective.

\section{HAServe: Scheduling-Memory Co-design}
\label{sec:design}
\label{sec:system_overview}

\begin{figure*}[!t]
\centering
\resizebox{0.85\linewidth}{!}{
\begingroup
\definecolor{huBlue}{HTML}{DCEEF5}
\definecolor{huGreen}{HTML}{E5F2EA}
\definecolor{huPink}{HTML}{F7DFDC}
\definecolor{huCream}{HTML}{FFF4DB}
\definecolor{huGray}{HTML}{F3F5F7}
\begin{tikzpicture}[x=1pt,y=1pt,
  font=\fontfamily{phv}\fontsize{8.5}{9.5}\selectfont,
  lab/.style={inner sep=0pt,align=center},
  title/.style={lab,font=\fontfamily{phv}\fontsize{9}{10}\selectfont\bfseries},
  box/.style={draw=black,line width=.8pt,rounded corners=2pt,fill=white},
  group/.style={box,dash pattern=on 3.5pt off 2.5pt},
  flow/.style={-{Stealth[length=3.7pt,width=3pt]},draw=black,line width=.8pt},
  update/.style={flow,dash pattern=on 3pt off 2pt,draw=black!65},
  token/.style={draw=black,line width=.55pt,fill=huGreen,
    minimum width=10pt,minimum height=11pt,inner sep=0pt},
  request/.style={draw=black,line width=.55pt,fill=white,
    minimum width=17pt,minimum height=11pt,inner sep=0pt},
  radix/.style={circle,draw=black,line width=.65pt,fill=white,
    minimum size=11pt,inner sep=0pt},
  eligible/.style={radix,fill=huPink,line width=.8pt},
  mandatory/.style={rectangle,double,draw=black,line width=.45pt,
    fill=huGray,minimum width=18pt,minimum height=12pt,inner sep=0pt}]
\path[use as bounding box] (0,0) rectangle (392,310);

\draw[box] (4,17) rectangle (308,306);
\node[title,fill=white,inner xsep=5pt] at (156,306) {HAServe};

\draw[box,fill=huBlue!35] (10,191) rectangle (78,289);
\node[title,font=\fontfamily{phv}\fontsize{8.5}{9.5}\selectfont\bfseries] at (44,277) {HorizonSchema};
\node[lab] at (29,263) {Age};
\node[lab] at (58,263) {Path};
\foreach \i/\y/\age in {1/248/13,2/230/10,3/212/6}{
  \node[lab] at (17,\y) {$q_{\i}$};
  \fill[black!60] (26,\y-1.5) rectangle +(\age,3);
  \ifnum\i<3
    \node[token] at (52,\y) {$A$};
    \node[token] at (64,\y) {$B$};
  \else
    \node[lab] at (58,\y) {short};
  \fi
}
\node[lab] at (44,198) {$\mathcal H_t$};
\draw[box,fill=huBlue!35] (10,148) rectangle (78,180);
\node[lab,font=\fontfamily{phv}\fontsize{8.5}{9.5}\selectfont\bfseries] at (44,171) {Runtime $s_t$};
\node[lab] at (44,158) {KV, refs, budgets};

\draw[group,fill=huBlue!25] (85,159) rectangle (299,289);
\node[title] at (192,281) {Horizon Scheduler};
\draw[box,fill=huBlue] (92,249) rectangle (292,271);
\node[lab,font=\fontfamily{phv}\fontsize{8.5}{9.5}\selectfont\bfseries] at (192,263) {Prefix organization};
\draw[black!55,line width=.45pt] (189,250)--(189,258);
\node[lab] at (139,254) {Original paths};
\node[lab] at (241,254) {Permitted alignment};
\draw[flow] (78,254)--(92,254);
\draw[flow] (150,249)--(150,243);
\draw[flow] (78,171)--(85,171);

\draw[box,fill=huCream] (92,165) rectangle (208,243);
\node[lab,font=\fontfamily{phv}\fontsize{8.5}{9.5}\selectfont\bfseries] at (150,232) {Candidate construction};
\draw[black,line width=.5pt,fill=huGreen] (98,205) rectangle (168,224);
\draw[black,line width=.5pt,fill=huBlue] (168,205) rectangle (202,224);
\node[lab] at (133,214.5) {Locality};
\node[lab] at (185,214.5) {FIFO\\reserve};
\node[lab] at (120,198) {$a_1$};
\node[lab] at (120,185) {$a_2$};
\node[request,fill=huGreen] at (149,198) {$q_1$};
\node[request,fill=huGreen] at (171,198) {$q_2$};
\node[request,fill=huGreen] at (149,185) {$q_1$};
\node[request] at (171,185) {$q_3$};
\draw[box] (98,166.5) rectangle (202,178);
\node[lab] at (150,172.25) {FIFO/LRU reference};

\draw[box,fill=huPink] (219,225) rectangle (292,243);
\node[lab,font=\fontfamily{phv}\fontsize{8.5}{9.5}\selectfont\bfseries] at (255.5,234) {Joint selection};
\draw[box,fill=huCream] (219,201) rectangle (292,217);
\node[lab] at (255.5,209) {Feasibility};
\draw[box,fill=huBlue] (219,165) rectangle (292,194);
\node[lab] at (255.5,179.5) {Work / allocation\\update};
\draw[flow] (255.5,194)--(255.5,201);
\draw[flow] (255.5,217)--(255.5,225);

\node[lab,anchor=west] at (153,152) {Paths, consumers};

\draw[box,fill=huGray!65] (10,28) rectangle (299,140);
\node[lab,font=\fontfamily{phv}\fontsize{8.5}{9.5}\selectfont\bfseries] at (107,131) {Private shadow evaluation};
\draw[flow] (44,148)--(44,140);
\draw[group,fill=huGreen!65] (16,31) rectangle (192,119);
\node[title] at (104,110) {DART};
\draw[box,fill=huCream] (45,91) rectangle (166,103);
\node[lab] at (105.5,97) {Consumer priorities};
\draw[flow] (55,91)--(55,86.5);

\node[radix] (root1) at (55,81) {};
\node[radix] (parent1) at (40,63) {$v$};
\node[mandatory] (ref1) at (76,63) {ref};
\node[eligible] (leaf1) at (40,51) {$u$};
\draw[black,line width=.65pt] (root1)--(parent1) (root1)--(ref1) (parent1)--(leaf1);
\node[radix] (root2) at (139,81) {};
\node[eligible] (parent2) at (123,63) {$v$};
\node[mandatory] (ref2) at (163,63) {ref};
\draw[black,line width=.65pt] (root2)--(parent2) (root2)--(ref2);
\draw[black!40,densely dashed,line width=.55pt] (123,57)--(123,54);
\draw[black!45,densely dashed,line width=.55pt] (123,49) circle (5.5pt);
\draw[black,line width=.7pt] (117,44)--(129,54);
\draw[flow] (87,72)--(109,72);
\node[lab] at (58,38) {Leaf removal};
\node[lab] at (146,38) {Parent recheck};

\draw[box,fill=huCream] (206,97) rectangle (290,116);
\node[lab] at (248,106.5) {Native LRU};
\draw[box,fill=huBlue] (206,45) rectangle (290,86);
\node[lab,font=\fontfamily{phv}\fontsize{8.5}{9.5}\selectfont\bfseries] at (248,76) {Exact rematch};
\draw[black,fill=huGreen,line width=.6pt] (215,59) rectangle (242,66);
\draw[black,densely dashed,line width=.6pt] (242,59) rectangle (281,66);
\node[lab] at (228.5,51.5) {Hit};
\node[lab] at (261.5,51.5) {Missing};
\draw[flow] (248,97)--(248,86);
\draw[flow] (192,68)--(206,68);
\draw[flow] (150,165)--(150,144)--(183,144)--(183,97)--(166,97);
\draw[flow] (150,144)--(249,144)--(249,116);
\draw[flow] (290,65)--(295,65)--(295,153)--(255.5,153)--(255.5,165);
\node[lab,fill=huGray!65,inner sep=1pt] at (274,133) {Matches};
\node[lab,fill=huGray!65,inner sep=1pt] at (274,122) {Capacity};

\draw[box,fill=huGray!60] (318,82) rectangle (391,244);
\node[title] at (354.5,234) {Backend};
\draw[box,fill=huPink] (325,201) rectangle (384,225);
\node[lab] at (354.5,213) {Validated\\commit};
\draw[flow] (292,234)--(312,234)--(312,213)--(325,213);
\draw[box,fill=huBlue] (325,154) rectangle (384,182);
\node[lab] at (354.5,168) {Prefill /\\decode};
\draw[flow] (354.5,201)--(354.5,182);
\draw[box,fill=huGreen] (325,100) rectangle (384,130);
\node[lab] at (354.5,115) {Live KV cache};
\draw[{Stealth[length=3.7pt,width=3pt]}-{Stealth[length=3.7pt,width=3pt]},draw=black,line width=.8pt] (354.5,154)--(354.5,130);

\draw[update] (354.5,82)--(354.5,6)--(0,6)--(0,234)--(10,234);
\draw[update] (0,164)--(10,164);
\node[lab,fill=white,inner sep=2pt] at (185,6) {Runtime events};
\end{tikzpicture}
\endgroup}

\caption{\textbf{HAServe Overview.} The scheduler compares feasible candidates on a shared snapshot by predicted mean remaining TTFT. Double borders mark mandatory references. Only the selected current action is committed.}
\Description{A unified HAServe architecture. HorizonSchema supplies request ages and exact paths. The same runtime snapshot supplies resident KV, references, and budgets to both the Horizon Scheduler and private shadow evaluation. Prefix organization feeds FIFO and locality-aware candidate construction. The locality-aware strategy includes a FIFO reserve. Alternative dispatches contain q1-q2 and q1-q3. Candidate paths, consumers, order, and retention settings enter private shadow evaluation. Consumer priorities guide DART reclamation, including eligible leaf removal and parent rechecks. Double borders mark mandatory references. Native LRU is a parallel retention branch. Both branches feed exact rematching. Returned matches and capacity feed work and allocation updates, feasibility checking, and joint selection in the scheduler. Only the selected current action enters validated backend commit. Prefill and decode read and update the live KV cache. Backend runtime events refresh request tracking and runtime state.}
\label{fig:haserve_execution}
\label{fig:prism_overview}
\label{fig:alignment_decision}
\label{fig:haserve_locality_batching}
\label{fig:haserve_dart_reclaim}
\end{figure*}

HAServe has two core components (Figure~\ref{fig:haserve_execution}). The \emph{Horizon Scheduler} tracks visible requests (Section~\ref{sec:visible_requests}), organizes prefixes (Section~\ref{sec:prefix_organization}), and constructs candidate batches (Section~\ref{sec:joint_selection}). The \emph{Demand-Aware Radix Tree} (DART) uses candidate consumers to evaluate retention and returns post-reclamation matches and capacity (Section~\ref{sec:kv_allocation}). The scheduler compares these joint choices and commits admission and retention together. Runtime events refresh both components (Section~\ref{sec:runtime_updates}).

\subsection{Tracking Visible Requests}
\label{sec:visible_requests}

On each arrival, HAServe uses $\mathcal T$ to establish the request's queue position and age, and $\mathcal P$ to register its token path and permitted serializations. These records associate visible requests with the prefixes they may consume. Requests sharing a segment identifier under different token ancestors remain consumers of different paths.

At each resource-feasible admission opportunity, the scheduler freezes the bounded pending set $\mathcal H_t$ and runtime snapshot $s_t$ defined in Section~\ref{sec:observable_horizon}. Candidates draw pending requests from $\mathcal H_t$ and are compared on this same set. Pending requests supply demand, while active references constrain compute and KV capacity. Only prefixes exposed as reusable by the backend contribute hits. Simultaneous consumers of a cold prefix incur separate missing work unless the backend provides shared computation.


\subsection{Prefix Organization}
\label{sec:prefix_organization}

The Horizon Scheduler's prefix-organization step establishes the legal paths and consumer groups shared by admission and retention. It turns HorizonSchema's prefix attributes $\mathcal P$ into a bounded set of serializations. Segment boundaries in $\Pi_i$ identify movable units, and $\mathcal L_i$ limits their permitted orders. The scheduler moves reusable segments toward the front while preserving tokens within segments, system instructions, and fixed history regions. Without segment annotations or a permitted alternative order, it groups original exact token paths without alignment. Fixed-order conversational requests such as ShareGPT take this branch, while scheduling and demand-aware retention remain active.

\noindent\textbf{Rank segments by visible demand.} For a segment $r$, let $g_t(r)$ count requests in $\mathcal Q_t$ containing it, $a_t(r)$ count requests in $\mathcal A_t$ containing it, and $n_t(r;\mathcal B)$ count its consumers in a provisional dispatch $\mathcal B$. The scheduler ranks segments using
\begin{equation}
\label{eq:segment_priority}
P_t(r;\mathcal B)=w_g g_t(r)+w_a a_t(r)+w_n n_t(r;\mathcal B).
\end{equation}
Here $w_g$, $w_a$, and $w_n$ weight queued, active, and provisional-dispatch demand. Alignment uses $\mathcal B=\emptyset$, while group scoring can use a provisional dispatch. Ordered signatures and arrival-ordered buckets reduce the request combinations considered (Algorithm~\ref{alg:joint_selection}). These counters guide grouping and later retention priorities. Exact token matches determine the resulting reuse.


\noindent\textbf{Check paths and form consumer groups.} The scheduler serializes each path, records its segment offsets, and checks its resident-prefix match and overlap with visible requests. Alignment can expose queue overlap while invalidating a cached prefix, so the original path $z_i$ remains an alternative. This step supplies the legal paths, token offsets, and arrival-ordered consumer groups for candidate construction in Section~\ref{sec:joint_selection}.

\noindent\textbf{Alignment trade-off.} A permitted order can expose reuse across service waves, but it can also invalidate resident-prefix matches and incur tokenization, serialization, and metadata work. These costs enter joint selection alongside capacity and waiting. The original path remains a reference choice (Section~\ref{sec:joint_selection}).

\subsection{Admission and Joint Selection}
\label{sec:horizon_scheduler}
\label{sec:joint_selection}

The Horizon Scheduler compares FIFO and locality-aware batches with candidate-specific retention. Algorithm~\ref{alg:joint_selection} gives the procedure; Appendix~\ref{sec:appendix:annotations} specifies group scores and construction details.

\noindent\textbf{Construct candidate batches.} Let $M$ be the current request-dispatch budget and $q_{\mathrm{cold}}$ the clipped FIFO reserve. The locality-aware strategy fills up to $M-q_{\mathrm{cold}}$ slots from ranked prefix groups and the reserve from the oldest remaining requests (Figure~\ref{fig:haserve_execution}). Each candidate carries its exact paths, consumer counts, and final token offsets. Group scores guide construction; Eq.~\eqref{eq:haserve_admission_cost} determines final selection.

\noindent\textbf{Compare joint serving choices.} HAServe evaluates a bounded set of dispatch, serialization, and retention choices $a$ on the same horizon. Each candidate pairs FIFO or locality-aware batching with LRU or DART retention and an associated budget. The scheduler evaluates reclamation under the selected retention rule, using DART for demand-aware retention or the native backend for LRU, and obtains updated prefix matches and available capacity (Section~\ref{sec:kv_allocation}). Only feasible candidates proceed to cost comparison. Returned matches and subsequent prefix construction determine $\widehat L_{i,t}^{a}$ and missing work in Eq.~\eqref{eq:visible_work}. Completed backend measurements supply service-time estimates conditioned on context length, batch composition, chunk limits, and active decode. The estimated mean remaining TTFT is
\begin{equation}
\label{eq:haserve_admission_cost}
\widehat J_t(a)=\widehat c_t(a)
+\frac{1}{|\mathcal H_t|}\sum_w N_{t,w}^{a}\widehat S_{t,w}^{a}
+\widehat{\overline T}^{\mathrm{1tok}}_t(a).
\end{equation}
Here $\widehat c_t(a)$ estimates shared blocking before the first prefill wave, $\widehat S_{t,w}^{a}$ estimates the duration of wave $w$, and $N_{t,w}^{a}$ counts requests in that wave or later. The term $\widehat{\overline T}^{\mathrm{1tok}}_t(a)$ estimates the mean interval from prefill completion to the first token, and $|\mathcal H_t|$ is the visible request count. This approximates the mean remaining latency in Eq.~\eqref{eq:horizon_area_identity} to rank candidates on the current horizon. Control and serialization costs enter the corresponding intervals, and retention-induced misses enter once through missing work. All choices use the same FIFO/LRU continuation to estimate the remaining horizon's cost. Only the current dispatch and retention decision are committed; selection repeats at the next admission opportunity.

\noindent\textbf{Relation to tail latency.} HAServe ranks candidates by predicted mean remaining TTFT over the visible horizon. Reducing repeated prefill and queue blocking may also shorten tail delays. The reported P99 gains are empirical results under the evaluated settings, not a guarantee of the objective.

\noindent\textbf{Balance locality with batch progress.} In Figure~\ref{fig:haserve_execution}, admitting $q_3$ with $q_1$ advances the short request but defers $q_2$'s reuse of $AB$. If $AB$ is cold, serving $q_1$ before $q_2$ may save duplicated prefix computation. Equation~\eqref{eq:haserve_admission_cost} weighs this work--waiting tradeoff after candidate-specific retention evaluation.

\noindent\textbf{Select and commit.} The age guard $W_{\max}$ gives old requests FIFO priority at the next resource-feasible opportunity. The chosen action must improve on FIFO/LRU by margin $\tau_t$. Throughout proposal construction and shadow evaluation, unsupported states, budget exhaustion, or snapshot changes invoke guarded full-queue FIFO/LRU. This fallback preserves guarded requests and fills feasible slots. Commit validates the state version atomically. Backend token, slot, chunk, and reference checks determine confirmed admissions. Appendix~\ref{sec:appendix:adaptive_design} specifies catalog settings and selection checks.

\subsection{KV Retention and Allocation}
\label{sec:dart}
\label{sec:kv_allocation}

DART evaluates demand-aware retention using each candidate's paths, consumers, and retention settings. It returns post-reclamation prefix matches and available capacity for the scheduler's work and feasibility checks. LRU candidates use the backend rule with the same physical-block accounting and rematching interface.

\noindent\textbf{Account for shared blocks.} Let $\mathcal K_{\mathrm{fixed}}$ contain blocks held by active references and other backend-mandatory state, $\mathcal K_{\mathrm{keep}}$ contain additional resident blocks preferred for retention, and $K_{\mathrm{new}}(\mathcal B)$ be the incremental allocation required by its next dispatch, including growth of active requests. With backend headroom $K_{\mathrm{reserve}}$ and physical pool size $K_{\mathrm{pool}}$, feasibility requires
\begin{equation}
\label{eq:kv_feasibility}
\left|\mathcal K_{\mathrm{fixed}}\cup\mathcal K_{\mathrm{keep}}\right|
+K_{\mathrm{new}}(\mathcal B)+K_{\mathrm{reserve}}
\le K_{\mathrm{pool}}.
\end{equation}
The retention footprint counts each physical block once, including shared ancestors. Incremental allocation follows the backend's block and chunk rules using each request's prefix match after simulated reclamation.

\noindent\textbf{Retain paths for their consumers.} Building on the backend radix cache~\citep{zheng24sglang}, DART maps each candidate serialization's token offsets to radix anchors marking exact reusable prefix boundaries. Dispatch consumer counts rank these anchors through the priority in Eq.~\eqref{eq:segment_priority}. Reclamation prefers private suffixes and lower-priority paths before preferred anchors, subject to mandatory references. Changing the candidate therefore changes both its consumers and the paths preferred for retention.

\noindent\textbf{Simulate reclamation and rematch.} For each DART candidate, the tree applies its retention priorities to a shadow copy of cache metadata, leaving the live cache unchanged. Reclamation follows eligible radix leaves, excludes mandatory references, and rechecks newly exposed parents. Preferred paths remain reclaimable when execution needs their space. DART then rematches the candidate's token paths against the resulting shadow cache and returns reusable-prefix lengths, available blocks, and any allocation shortfall. The scheduler recomputes missing work and incremental allocation from these matches, avoiding the initial-match ranking error in Appendix~\ref{sec:appendix:ranking_reversal}. If the recomputed allocation exceeds available capacity or violates Eq.~\eqref{eq:kv_feasibility}, the candidate is discarded before cost comparison.

\subsection{Runtime Updates}
\label{sec:runtime_updates}
\label{sec:joint_execution}
\label{sec:control_overhead}

Arrival, admission, prefill, completion, and reclamation events update demand, references, and residency. Backend-confirmed admissions determine final token offsets and consumer counters for the actual admitted subset. Newly released successors register their own exact-path demand. Completed timing events supply calibration and prediction checks for the cost model.

\noindent\textbf{Control cost.} For $N_{\mathrm{seg}}$ visible segment occurrences and $N_{\mathrm{grp}}$ groups, accounting and grouping take expected $O(N_{\mathrm{seg}})$ work, with $O(N_{\mathrm{grp}}\log N_{\mathrm{grp}})$ group ranking. With $V$ resident metadata nodes, a full eligible-leaf heap refresh costs $O(V)$ and reclaiming $E$ nodes costs $O(E\log V)$. Timing also includes tokenization, matching, serialization, block accounting, shadow evaluation, and commit overhead. Measured control-layer costs are reported in Section~\ref{sec:eval:control_overhead}.

\subsection{Overall Scheduling Algorithm}
\label{sec:algorithm_overview}

Algorithm~\ref{alg:joint_selection} summarizes candidate construction, retention evaluation, and joint selection. The finite catalog $\Theta$ specifies grouping, FIFO reserve, and retention settings; Appendix~\ref{sec:appendix:algorithms} provides their definitions. The output is the backend-confirmed dispatch and its retention settings.

\begin{breakablealgorithm}{HAServe's online scheduling and retention algorithm.}{alg:joint_selection}
\begin{algorithmic}
\Statex \textbf{Input:} Frozen horizon $\mathcal H_t$, runtime state $s_t$, catalog $\Theta$, age guard $W_{\max}$, benefit margin $\tau_t$, evaluation budget
\Statex \textbf{Output:} Backend-confirmed admissions and joint retention settings
\Statex \textbf{Guard:} At any step, unsupported states, budget exhaustion, or snapshot changes invoke fresh full-queue FIFO/LRU with the age guard
\State \parbox[t]{\linewidth}{\textbf{Step 1: Construct legal dispatch candidates}\\
Read queued/active counters from $s_t$ and derive dispatch budget $M$}
\State Construct next-feasible original-path FIFO/LRU reference $a_t^0$ with the age guard
\State $\mathcal X_t\gets\{a_t^0\}$
\ForAll{configuration $\theta\in\Theta$}
  \State Choose permitted or original paths for $\mathcal H_t$ and compute token offsets
  \State Build arrival-ordered groups from ordered signatures or exact paths
  \State Match initial prefixes and clip $q_{\mathrm{cold}}$ to $M$
  \If{$\theta$ selects locality-aware batching}
    \State Score groups using their oldest up to $M-q_{\mathrm{cold}}$ consumers (Eq.~\eqref{eq:bucket_score})
    \State Fill $\mathcal B$ with up to $M-q_{\mathrm{cold}}$ requests by score, FIFO within groups
    \State Append up to $q_{\mathrm{cold}}$ oldest unselected requests from $\mathcal H_t$
  \Else
    \State Form FIFO batch $\mathcal B$ of up to $M$ requests using the chosen paths
  \EndIf
  \State Apply $W_{\max}$ and refresh paths, offsets, consumer counts, and priorities
  \State Form $a$ from $\mathcal B$ and $\theta$'s retention settings $(e_a,K_{\mathrm{prot},a})$
  \State Add $a$ to $\mathcal X_t$ if its order, paths, and retention settings are distinct
\EndFor
\State \parbox[t]{\linewidth}{\textbf{Step 2: Evaluate admission and retention jointly}\\
$\mathcal F_t\gets\emptyset$}
\ForAll{$a\in\mathcal X_t$, starting with $a_t^0$}
  \State $(\widetilde s_a,K_{\mathrm{avail}},\mathrm{shortfall})\gets\textsc{ShadowReclaim}(s_t,a)$
  \State Rematch exact paths against the private shadow state $\widetilde s_a$
  \State Recompute missing work and incremental allocation $K_{\mathrm{new}}(\mathcal B_a)$
  \If{no shortfall, $K_{\mathrm{new}}(\mathcal B_a)\le K_{\mathrm{avail}}$, and backend checks pass}
    \State Estimate $\widehat J_t(a)$ with common FIFO/LRU continuation (Eq.~\eqref{eq:haserve_admission_cost})
    \State $\mathcal F_t\gets\mathcal F_t\cup\{a\}$
  \ElsIf{$a=a_t^0$}
    \State \Return fresh full-queue FIFO/LRU with the age guard
  \EndIf
\EndFor
\State \parbox[t]{\linewidth}{\textbf{Step 3: Select and commit the current action}\\
$a^\star\gets\arg\min_{a\in\mathcal F_t}\widehat J_t(a)$}
\State \algorithmicif\ $\widehat J_t(a_t^0)-\widehat J_t(a^\star)\le\tau_t$ \algorithmicthen\ $a^\star\gets a_t^0$
\State Validate the state version and atomically commit only $a^\star$
\State Refresh metadata from backend-confirmed admissions
\State \Return confirmed admissions
\end{algorithmic}
\end{breakablealgorithm}
\FloatBarrier

\section{Evaluation}
\label{sec:evaluation}

We describe the workloads and arrival process in Section~\ref{sec:eval:workload}, then evaluate end-to-end serving performance in Section~\ref{sec:eval:performance}. Component ablations follow in Section~\ref{sec:eval:ablation}, with answer quality and control overhead in Section~\ref{sec:eval:quality_overhead}. The controlled observations in Section~\ref{sec:motivation_observations} establish the serving trade-offs behind these comparisons.

\subsection{Experimental Setup}

\label{sec:eval:setup}

\noindent\textbf{System Implementation.} HAServe is implemented on SGLang~\citep{zheng24sglang} with the original continuous batching and chunked-prefill mechanisms retained. The Horizon Scheduler organizes visible requests and selects joint serving choices on top of SGLang, while DART extends the backend radix tree with consumer-aware retention.

\noindent\textbf{Hardware and Inference Engines.} The updated evaluation uses eight NVIDIA H100 GPUs. We use SGLang v0.5.21 and vLLM v0.31.0~\citep{kwon23vllm}, with each engine version held fixed across its associated comparisons.

Each baseline runs on the inference engine underlying its implementation: SGLang for k-LPM and ContextPilot, and vLLM for CacheBlend and UniCache. This selection provides broad coverage across two inference engines and four complementary optimization layers: request scheduling, context organization, cached-context fusion, and cache eviction.

\noindent\textbf{Baselines.}
The updated comparison set contains four baselines:
\textbf{k-LPM}~\citep{dexter25klpm}, \textbf{ContextPilot}~\citep{contextpilot2026}, \textbf{CacheBlend}~\citep{yao25cacheblend}, and \textbf{UniCache}~\citep{ouyang26unicache}. Comparisons use the same model, materialized requests, arrival timestamps, concurrency limit, and GPU-memory budget; methods built on the same engine share its version and common runtime settings.
\begin{itemize}[leftmargin=*,topsep=2pt,itemsep=2pt]
  \item \textbf{k-LPM}: k-LPM enhances SGLang's longest-prefix-match scheduling by balancing prefix reuse with latency constraints, making it a direct comparison with the locality-aware batching strategy of the Horizon Scheduler. We use $k=2$, which achieved the lowest P99 TTFT across most evaluated request rates in the original study.
  \item \textbf{ContextPilot}: ContextPilot runs on SGLang and uses a context index, alignment, and deduplication to improve reuse, providing a comparison with optimization at the context-management layer.
  \item \textbf{CacheBlend}: CacheBlend runs on vLLM and combines cached KV states for multiple context chunks with selective recomputation, providing a comparison with cached-context fusion.
  \item \textbf{UniCache}: UniCache runs on vLLM and combines session and structural reuse priorities with dynamic cache allocation across tasks, providing a comparison with workload-aware prefix-cache eviction.
\end{itemize}

\noindent\textbf{Models.} We evaluate \textbf{Qwen3-4B-Instruct-2507}~\citep{qwen3} and the \textbf{Llama2-13B} base model~\citep{touvron23llama2}, representing compact and memory-intensive serving regimes. Llama2-13B uses BF16, tensor parallelism across all eight GPUs, and a 32{,}768-token context with linear RoPE scaling by a factor of eight.

\noindent\textbf{Repeated measurements.} We measure each experimental setting five times using different random seeds and report the arithmetic mean of the five measurements.

\noindent\textbf{Metrics.} 
We evaluate the performance and accuracy by: 
\begin{itemize}[leftmargin=*,topsep=2pt,itemsep=2pt]
\item \textbf{Latency.} We report P50, P90, P95, and P99 TTFT to characterize both typical serving efficiency and long-tail latency.
\item \textbf{Throughput.} We report throughput as the number of completed requests per second. This measures the aggregate serving rate under each offered load and complements TTFT, which captures per-request responsiveness.
\item \textbf{KV-Cache reuse.} We report KV-cache hit rate to measure realized reuse. Qwen3-4B uses backend exact-prefix hit rate. Llama2-13B reports actually reused input tokens divided by all input tokens, including the reused ranges reported by CacheBlend.
\item \textbf{Answer quality.} We report answer F1 on question-answering workloads as a correctness guardrail (Section~\ref{sec:eval:quality_overhead}). The $\tau$-bench trace evaluates serving performance rather than agent task success.
\end{itemize}

\subsection{Workloads and Arrival Process}
\label{sec:eval:workload}

We use five benchmarks: MultiHopRAG, ShareGPT, LooGLE, $\tau$-bench, and a mixed workload combining requests from all four. All request streams use Zipfian sampling with exponent $\alpha=1.2$ to model skewed popularity and context recurrence, while Poisson arrivals control request timing. We use arrival rates of 100, 200, and 400 QPS for Qwen3-4B, and 50, 75, and 100 QPS for Llama2-13B. These rates stress both models under fixed budgets to expose queueing, repeated prefill, and KV-cache contention. We generate timestamps once per trace and replay the same materialized prompts and timestamps across methods.

\noindent\textbf{MultiHopRAG.}
MultiHopRAG~\citep{tang24multihoprag} supplies multi-hop question-answering requests. Each prompt combines a system prefix, retrieved evidence, and a request-specific suffix, exposing reuse across questions with overlapping evidence.

\noindent\textbf{ShareGPT.}
ShareGPT~\citep{sharegpt2023} supplies conversational requests. We serialize each request with its preceding dialogue history to expose prefix reuse across turns in the same conversation.

\noindent\textbf{LooGLE.}
LooGLE~\citep{li24loogle} supplies long-context question-answering requests. Each prompt contains a complete document and a question, exposing reuse across questions over the same document.

\noindent\textbf{$\tau$-bench.}
We use the $\tau$-bench-derived AgentPrefix trace~\citep{yao24taubench}, preserving policies, tool definitions, task records, dialogue history, and deterministic tool outputs. These shared and evolving prompt regions expose prefix reuse across agent requests. The measurements evaluate prompt serving rather than agent task success.

\noindent\textbf{Mixed.}
We combine requests from MultiHopRAG, ShareGPT, LooGLE, and $\tau$-bench into continuous traces. The four request types contribute approximately equal total token counts, while seed-dependent sampling produces substantial variation in their proportions across time intervals. The changing composition alters reusable context while the cache carries over, exercising online adaptation to changes in request mix and reuse patterns. The results below summarize each full trace.

\subsection{End-to-End Performance}
\label{sec:eval:performance}

TTFT is measured in seconds, cache hit rate in percent, and throughput in completed requests per second. Relative improvements use the strongest of the four baselines for the same model, workload, load setting, and metric.

\noindent\textbf{TTFT across workloads.}
On the mixed workload, P99 TTFT decreases by 16.4--53.6\% for Qwen3-4B and 8.6--22.6\% for Llama2-13B. Across all fifteen workload--load settings, Llama2-13B P99 improves by 1.1--22.6\%.

\noindent\textbf{Cache reuse and throughput.}
On the mixed workload, throughput increases by 9.3--36.0\% for Qwen3-4B and 15.7--18.7\% for Llama2-13B, while hit rate exceeds the highest-hit baseline by 4.28--9.70 and 5.99--9.55 percentage points, respectively.

\begin{figure*}[!t]
{\raggedright \noindent\textbf{Qwen3-4B results.} Figures~\ref{fig:eval:ttft_matrix} and~\ref{fig:eval:reuse_throughput} report latency, reuse, and throughput across five workloads at 100, 200, and 400 QPS, using eight H100 GPUs with TP=8.\par}
\smallskip
\centering
\includegraphics[width=0.85\linewidth]{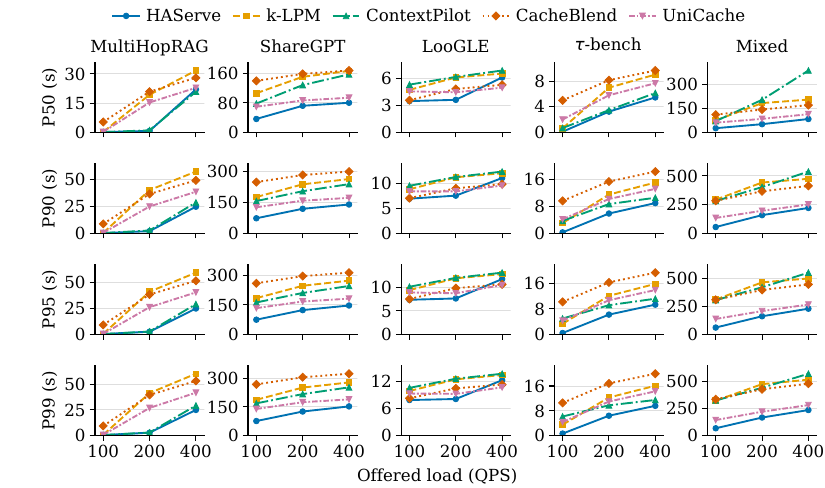}
\caption{\textbf{Qwen3-4B TTFT across five workloads.} Rows show P50, P90, P95, and P99 in seconds. All points are means over five random seeds.}
\Description{Twenty line-chart panels show four TTFT percentiles across five workloads, with one shared method legend.}
\label{fig:eval:ttft_matrix}
\end{figure*}

\begin{figure*}[!t]
\centering
\includegraphics[width=0.85\linewidth]{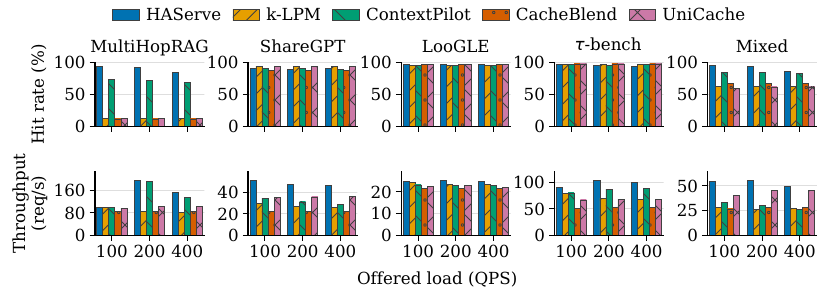}
\caption{\textbf{Qwen3-4B cache reuse and throughput.} Top: exact-prefix hit rate. Bottom: completed requests per second.}
\Description{Ten grouped bar panels show hit rate and throughput across five workloads and three request rates.}
\label{fig:eval:reuse_throughput}
\end{figure*}

\begin{figure*}[!t]
{\raggedright \noindent\textbf{Llama2-13B results.} Figures~\ref{fig:eval:llama_ttft_matrix} and~\ref{fig:eval:llama_reuse_throughput} report the corresponding metrics at 50, 75, and 100 QPS, using BF16 and eight H100 GPUs with TP=8.\par}
\smallskip
\centering
\includegraphics[width=0.85\linewidth]{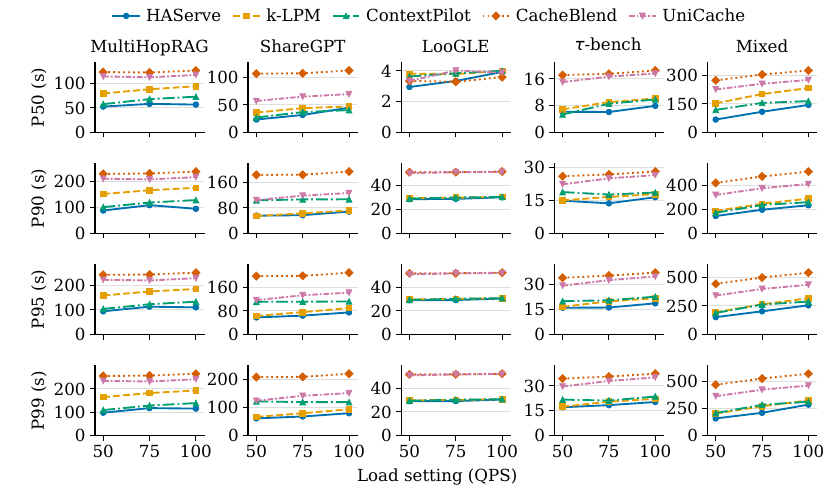}
\caption{\textbf{Llama2-13B TTFT across five workloads.} Rows show P50, P90, P95, and P99 in seconds. All points are means over five random seeds.}
\Description{Twenty line-chart panels show four TTFT percentiles across five workloads for Llama2-13B. Five methods use the same colors, markers, and line styles as the Qwen3-4B figure. Arrival rates are 50, 75, and 100 QPS for all workloads.}
\label{fig:eval:llama_ttft_matrix}
\end{figure*}

\begin{figure*}[!t]
\centering
\includegraphics[width=0.85\linewidth]{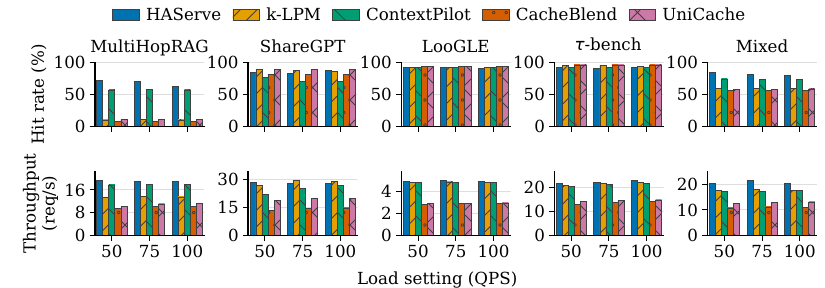}
\caption{\textbf{Llama2-13B cache reuse and throughput.} Top: reused input tokens as a percentage of all input tokens. Bottom: completed requests per second.}
\Description{Ten grouped bar panels show cache hit rate and throughput across five workloads and three load settings for Llama2-13B. Reported values are means over five runs with different random seeds.}
\label{fig:eval:llama_reuse_throughput}
\end{figure*}

\noindent\textbf{Benefits across workloads and load levels.}
At 100 QPS, Qwen3-4B P99 decreases by 45.7\% on ShareGPT and 83.0\% on the $\tau$-bench trace. These gains are consistent with serving related consumers together and retaining shared histories or policy and tool prefixes for pending requests.

For Qwen3-4B on MultiHopRAG, P99 reduction grows from 2.9\% at 100 QPS to 14.7\% at 400 QPS, where throughput also improves by 12.8\%. This pattern is consistent with avoiding repeated prefill becoming more valuable as requests compete for service. HAServe weighs saved work against waiting, while DART retains paths needed by pending consumers. On Mixed at the highest tested rates, Qwen3-4B retains a 16.4\% P99 reduction and a 9.3\% throughput gain, while Llama2-13B achieves 8.6\% and 15.7\%, respectively.

Appendix~\ref{sec:appendix:multi_replica} reports additional mixed-workload results with two TP=4 replicas on the same eight-GPU deployment.

\subsection{Component Ablations}
\label{sec:eval:ablation}

\begin{figure*}[!t]
\centering
\includegraphics[width=0.85\linewidth]{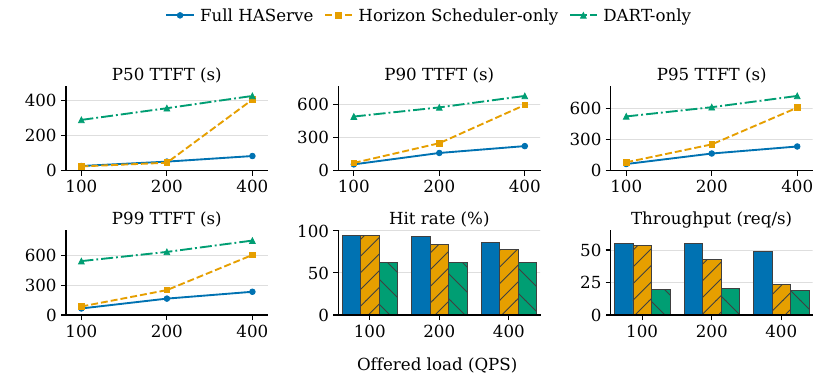}
\caption{\textbf{Qwen3-4B-Instruct-2507 component ablations on the continuous mixed workload.} Four TTFT percentiles, exact-prefix hit rate, and throughput compare the full system with its two component configurations at 100, 200, and 400 QPS. Values are means over five random seeds.}
\Description{Six panels compare Full HAServe, Horizon Scheduler-only, and DART-only using line charts for TTFT and grouped bars for hit rate and throughput.}
\label{fig:eval:ablation_matrix}
\end{figure*}

\noindent\textbf{Component configurations.}
Figure~\ref{fig:eval:ablation_matrix} reports three Qwen3-4B-Instruct-2507 component configurations on the continuous mixed workload. Full HAServe includes both the Horizon Scheduler and DART. Horizon Scheduler-only retains the scheduler and uses LRU retention. DART-only retains demand-aware retention with FIFO admission. The Full HAServe series is shared with the main mixed-workload comparison.

\noindent\textbf{Component contributions.}
Full HAServe achieves lower P99 TTFT and higher throughput than either component configuration at all three rates. At 400 QPS, it reduces P99 by 61.3\% and increases hit rate by 8.90 percentage points relative to Horizon Scheduler-only. Its P99 is also 68.8\% lower than DART-only. FIFO can separate related consumers, so DART-only may hold reusable prefixes before their next use, occupying capacity while reuse is delayed. Joint selection aligns retention with consumer timing.

\subsection{Answer Quality and Control Overhead}
\label{sec:eval:quality_overhead}

\noindent\textbf{Answer quality.}
\label{sec:eval:answer_quality}
We compare Qwen3-4B-Instruct-2507 answer F1 across HAServe and the four baselines using identical questions and evidence. Quality measurements use the eight-H100 configuration and the same materialized prompts as the corresponding serving comparisons.

\begin{figure*}[!t]
\centering
\includegraphics[width=0.85\linewidth]{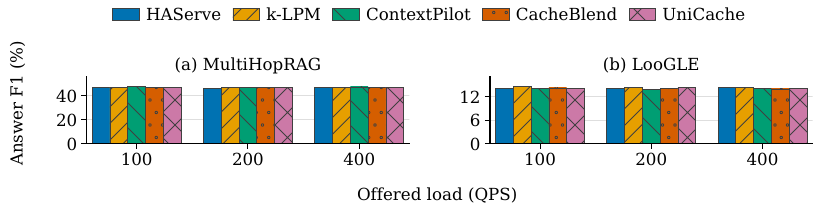}
\caption{\textbf{Answer F1 on MultiHopRAG and LooGLE.} Grouped bars show Qwen3-4B-Instruct-2507 results at 100, 200, and 400 QPS on eight H100 GPUs, averaged over five random seeds.}
\Description{Two grouped bar panels compare answer F1 for five methods at three request rates.}
\label{fig:eval:answer_f1}
\end{figure*}

Figure~\ref{fig:eval:answer_f1} reports Qwen3-4B-Instruct-2507 answer F1 on MultiHopRAG and LooGLE at each request rate. Appendix~\ref{sec:appendix:organization_quality} discusses the scope of these quality measurements. Across matched request rates, HAServe's absolute F1 gap from any baseline is at most 1.16 percentage points on MultiHopRAG and 0.32 on LooGLE. Thus, the serving gains are accompanied by small changes in answer quality under these QA settings.

\noindent\textbf{Control overhead.}
\phantomsection
\label{sec:eval:control_overhead}
Table~\ref{tab:eval:control_overhead} reports HAServe control-layer latency percentiles and P99 control/TTFT ratios on the Qwen3-4B mixed workload at 100, 200, and 400 QPS. Median overhead increases from 6.21\,ms at 100 QPS to 18.08\,ms at 400 QPS, while P99 overhead increases from 15.62\,ms to 53.51\,ms. On the low-latency MultiHopRAG workload at 100 QPS, mean control overhead is 2.34\,ms, with P50, P95, and P99 of 1.53, 6.71, and 12.02\,ms, respectively. The P50, P95, and P99 control overheads are approximately 2.19\%, 2.49\%, and 3.54\% of the corresponding TTFT values of 70, 270, and 340\,ms, respectively.

\begin{table}[htbp]
\centering
\caption{HAServe control-layer overhead on the Qwen3-4B mixed workload (five-seed means), with P99 control/TTFT ratios.}
\label{tab:eval:control_overhead}
\small
\setlength{\tabcolsep}{3pt}
\begin{tabular}{@{}rrrrrr@{}}
\toprule
QPS & P50 (ms) & P90 (ms) & P95 (ms) & P99 (ms) & \shortstack{P99 control/\\P99 TTFT (\%)} \\
\midrule
100 & 6.21 & 10.15 & 12.22 & 15.62 & 0.024 \\
200 & 12.63 & 23.02 & 25.94 & 39.05 & 0.024 \\
400 & 18.08 & 28.91 & 32.24 & 53.51 & 0.023 \\
\bottomrule
\end{tabular}
\end{table}

\section{Related Work}
\label{sec:related}

\noindent\textbf{Structured workloads and serving interfaces.}
Serving interfaces expose reusable structure or application dependencies. Prompt Cache~\citep{gim24promptcache} specifies prompt modules and positions for modular attention reuse. Parrot~\citep{lin24parrot} exposes dependencies through Semantic Variables, while Ayo~\citep{tan25ayo} represents LLM and non-LLM operations in a dataflow graph.

Workflow-aware systems use application dependencies to guide online decisions. KVFlow~\citep{pan25kvflow} uses an Agent Step Graph for eviction and prefetching, together with loading-status-aware scheduling. TOPAS~\citep{ni26topas} jointly selects resident agent prefixes and request admission under a shared KV budget, using a known workflow DAG to target task-level completion time. HAServe senses demand online from arrived requests and runtime cache state. HorizonSchema exposes their exact prefix paths and visible consumers, enabling joint admission and retention for request TTFT without requiring a workflow graph or counting unreleased successors as visible demand.

\noindent\textbf{LLM serving and scheduling.}
Orca~\citep{yu22orca}, vLLM~\citep{kwon23vllm}, and SGLang~\citep{zheng24sglang} provide iteration-level scheduling, paged KV memory, and radix-prefix caching. FastServe~\citep{wu23fastserve}, FlexGen~\citep{sheng23flexgen}, Sarathi-Serve~\citep{agrawal24sarathi}, and DistServe~\citep{zhong24distserve} improve execution through preemption, offloading, chunked prefill, or phase separation. Scheduling also targets decode deadlines~\citep{bari25slai} and client fairness~\citep{sheng24vtc,cao25dlpm}. Among reuse-oriented approaches, k-LPM~\citep{dexter25klpm} balances prefix locality with latency constraints, while ContextPilot~\citep{contextpilot2026} combines context indexing, alignment, deduplication, and cache-aware scheduling.

Across instances, Preble~\citep{srivatsa25preble} balances reuse and load, including eviction-induced recomputation in placement costs. LMetric~\citep{zhang26lmetric}, Llumnix~\citep{sun24llumnix}, and SageServe~\citep{jaiswal25sageserve} address routing, migration, or placement. HAServe jointly evaluates admission and retention within an instance using a finite-horizon TTFT estimate of saved work and imposed waiting.

\noindent\textbf{Context reuse and cache management.}
PipeRAG~\citep{jiang24piperag} and RAGO~\citep{jiang25rago} optimize RAG pipelines and resource use. RAGCache~\citep{jin2025ragcache} manages reusable knowledge states, CacheBlend~\citep{yao25cacheblend} selectively recomputes cached context, and TurboRAG~\citep{lu2025turborag} reuses precomputed chunks. Other systems couple prefix-aware execution with attention kernels~\citep{ye24chunkattention}, compress KV states~\citep{liu24cachegen}, or combine selective recomputation with scheduling~\citep{yang25kvshare}.

Retention policies use prefix hotness~\citep{li25hotprefix}, session and structural reuse with dynamic capacity allocation~\citep{ouyang26unicache}, or token roles and inter-turn timing~\citep{fang26saecache}. HAServe prioritizes exact-prefix paths by candidate consumers and checks post-reclamation matches and backend allocation constraints before selecting admission. It restricts prompt alignment to application-permitted orders and evaluates answer quality separately.

\section{Conclusion and Future Outlook}
\label{sec:conclusion}

This work connects workload characterization to scheduling--memory co-design for online LLM serving. HorizonSchema represents exact prefix paths, visible consumers, and arrival timing, revealing how prefix survival and service order shape realized reuse and latency. HAServe turns these insights into joint admission and retention through the Horizon Scheduler and DART, balancing post-reclamation reuse and capacity against queue-wide waiting. Our SGLang implementation achieves up to $5.90\times$ P99 TTFT speedup and up to $1.44\times$ the throughput of the strongest baseline for each metric and setting across two models and five workloads, measured over five seeds on eight H100 GPUs.

Looking ahead, the path--consumer abstraction provides a basis for coordinating request placement and prefix residency across replicas, accounting for queueing and transfer costs. It could also guide KV placement across GPU and host memory, choosing which prefixes to retain, move, or recompute. These extensions would apply the same design principle: evaluate reuse through saved work, occupied capacity, and waiting imposed across visible requests.

\FloatBarrier

\bibliographystyle{ACM-Reference-Format}
\bibliography{references}

\appendix
\section{Implementation Details}
\label{sec:appendix:algorithms}

This appendix supplements Algorithm~\ref{alg:joint_selection} in Section~\ref{sec:algorithm_overview} with group scores, retention metadata, horizon settings, and selection checks.

\subsection{Locality-Aware Batch Construction}
\label{sec:appendix:annotations}

The candidate-construction step in Algorithm~\ref{alg:joint_selection} takes the frozen horizon $\mathcal H_t$, runtime snapshot $s_t$, and one configuration $\theta$ from Eq.~\eqref{eq:controller_config}. Only requests in $\mathcal H_t$ enter its candidate batch. The snapshot supplies queued counters over the full pending set $\mathcal Q_t$ and active counters over $\mathcal A_t$. The dispatch budget $M$ follows the backend's current slot, token, and chunk limits. It is split between locality-ranked requests and a FIFO reserve of up to $q_{\mathrm{cold}}$ requests, with quotas clipped to the current budget.

\textsc{LegalPath} moves high-priority reusable segments toward the front within application-permitted regions and serializes the result. It preserves immutable regions, tokens within segments, and the relative order of unselected segments. \textsc{OrderedSignature} selects up to $\kappa$ segments by $P_t(r;\emptyset)$ and retains their order in the resulting path. For requests without segment annotations, \textsc{ExactPathGroup} groups the original token paths without alignment. Group keys only guide candidate construction. Exact token matches determine reuse, as described in Section~\ref{sec:prefix_organization}.

The queued, active, and candidate-dispatch segment counters are $g_t(r)$, $a_t(r)$, and $n_t(r;\mathcal B)$ for segment $r$ and reference batch $\mathcal B$. Their priority $P_t(r;\mathcal B)$ follows Eq.~\eqref{eq:segment_priority}. The reported fixed settings use counter weights $(w_g,w_a,w_n)=(1,10^6,10^5)$. Let $\mathcal G[b]$ be the arrival-ordered requests in bucket $b$, and let $\mathcal R_b$ be the reusable segments represented by that bucket. Its provisional hot batch $\mathcal B_b^{\mathrm{hot}}$ contains up to $M-q_{\mathrm{cold}}$ requests. The bucket utility and score are
\begin{equation}
\label{eq:bucket_score}
\begin{aligned}
U(b)&=\frac{\sum_{r\in\mathcal R_b}P_t(r;\mathcal B_b^{\mathrm{hot}})}{|\mathcal R_b|},\\
\operatorname{Score}(b)&=\alpha_{\mathrm{size}}|\mathcal G[b]|+\beta_{\mathrm{util}}U(b),
\end{aligned}
\end{equation}
where $\alpha_{\mathrm{size}}$ and $\beta_{\mathrm{util}}$ weight request multiplicity and mean segment utility. An empty segment set has zero utility. Each bucket's provisional counters remain fixed during the final score-ranked scan. The constructed batch has its own recomputed dispatch counters and priorities. Its metadata contains request IDs, bucket positions, exact token paths, and final segment offsets. All updates are local to the candidate. Algorithm~\ref{alg:joint_selection} applies the common age guard before retention evaluation, and backend-confirmed updates follow Section~\ref{sec:runtime_updates}.

\subsection{Retention Metadata and Reclamation}

DART associates reusable anchors with exact radix paths using the candidate serialization's token offsets. The eviction ordering reclaims private suffixes before low-priority reusable nodes and preferred anchors, subject to active-reference constraints. An anchor preference does not override mandatory references or the capacity needed for execution.

For candidate evaluation, $K_{\mathrm{prot}}$ caps the shared footprint of preferred paths in KV blocks. Ancestors required by retained paths count toward that budget, and shared blocks count once. \textsc{ShadowReclaim} uses the candidate's consumers and retention settings $(e,K_{\mathrm{prot}})$ to simulate reclamation on a private copy of $s_t$. It follows DART priorities when $e$ selects DART and the native backend rule when $e$ selects LRU. Given a reclaim target $K_{\mathrm{free}}$, it returns the resulting shadow state $\widetilde s_a$, the blocks $K_{\mathrm{avail}}$ available for incremental allocation after reserving backend headroom, and any reclamation shortfall.

The selector rematches exact paths against $\widetilde s_a$ before recomputing missing work and $K_{\mathrm{new}}(\mathcal B_a)$ for candidate batch $\mathcal B_a$. A candidate passes the post-rematch check only if there is no shortfall, $K_{\mathrm{new}}(\mathcal B_a)\le K_{\mathrm{avail}}$, Eq.~\eqref{eq:kv_feasibility} holds, and the backend's token, slot, chunk, and reference constraints are satisfied. A failed check discards the candidate before cost comparison.

Reclamation refreshes the eligible-leaf priority heap after a global demand or retention-mode change. Checking only the next popped entry cannot discover a decreased key elsewhere in the heap. Each removed leaf must still be attached and unreferenced, and a newly eligible parent receives a priority from the same frozen decision state. A global update during reclamation invalidates the evaluation. Section~\ref{sec:control_overhead} includes heap-refresh cost separately from per-node reclamation.

\subsection{Joint-Selection Settings}
\label{sec:appendix:adaptive_design}
\label{sec:controller_state}

\noindent\textbf{Horizon settings.} Let $t_j$ be observation epoch $j$ and $A_j$ the number of arrivals since $t_{j-1}$. For a positive observation interval, the smoothed arrival rate is
\begin{equation}
\label{eq:arrival_ewma}
\widehat\lambda_j=(1-\gamma)\widehat\lambda_{j-1}
+\gamma\frac{A_j}{t_j-t_{j-1}},\qquad 0<\gamma\leq1,
\end{equation}
where $\gamma$ is the smoothing weight. The selector freezes the oldest $H_t$ pending requests with
\begin{equation}
\label{eq:controller_horizon_size}
H_t=\min\!\left\{|\mathcal Q_t|,H_{\max},
\max\{H_{\min},\lceil\widehat\lambda_t T_{\mathrm{look}}\rceil\}\right\}.
\end{equation}
Here $\widehat\lambda_t$ is the latest rate estimate, $H_{\min}$ and $H_{\max}$ are horizon-size limits, and $T_{\mathrm{look}}$ is an inspection timescale. These settings are fixed using calibration traces. They bound inspection of already visible requests and add no collection wait. Full-queue FIFO bypasses scoring if the bound would omit a guarded oldest request or unnecessarily leave feasible slots idle.

\noindent\textbf{Candidate settings.} A finite catalog $\Theta$ contains configurations
\begin{equation}
\label{eq:controller_config}
\theta=(o,\kappa,q_{\mathrm{cold}},e,K_{\mathrm{prot}}),
\end{equation}
where $o$ selects FIFO or locality-aware batching within the Horizon Scheduler and $e$ selects LRU or DART retention. The remaining settings control grouping, the FIFO reserve, and the protected footprint. The catalog includes FIFO/LRU, locality-aware/LRU, and FIFO or locality-aware batching with bounded DART retention. The FIFO/LRU reference preserves original paths as an alternative to aligned candidates. Each candidate contains at most one serialization of each request. Setting $K_{\mathrm{prot}}=0$ removes extra protection but does not make DART equivalent to LRU. Illegal transformations are masked, quotas are clipped to feasible capacity, and duplicate realized actions are removed. Counter weights are not searched online.

\noindent\textbf{Common continuation and calibration.}
\label{sec:controller_selection}
Each configuration proposes one dispatch and its associated reclamation from the same state snapshot. The FIFO/LRU reference and all candidates use FIFO/LRU on the remaining horizon, with no further prompt rewriting. A prefix becomes reusable only after its construction completes under backend rules. The service model uses missing tokens, existing context lengths, batch composition, chunk limits, and active decode. Control work, candidate tokenization, and metadata updates are included in the cost even when selection falls back. Appendix~\ref{sec:appendix:decision_cost} gives the accounting.

An admission-age threshold $W_{\max}$ gives guarded requests FIFO priority at the next resource-feasible opportunity in every candidate and fallback. It bounds discretionary overtaking. A candidate is selected when its estimated benefit over the next-feasible FIFO/LRU action $a_t^0$ exceeds margin $\tau_t$, calibrated using held-out paired comparisons.

The next-feasible FIFO/LRU reference $a_t^0$ preserves original paths and is evaluated first through the same reclamation, rematching, and scoring steps. Proposal identity includes dispatch order, exact paths, and retention settings, so a shared batch alone does not imply a duplicate. Proposal construction and evaluation may be interleaved in catalog order to keep control work bounded.

\noindent\textbf{Fallback rule.} Throughout construction and evaluation, an unsupported state, an exhausted budget, or a changed state version immediately returns control to fresh full-queue FIFO/LRU with the common age guard. Checks also run inside each shadow evaluation, including the reference. The same fallback applies if the horizon would omit a guarded request or leave feasible slots idle. Shadow evaluation never changes the live cache. Commit validates the state version atomically and applies only the selected action's required reclamation and serving settings.

\section{Visible-Demand Cost Accounting}
\label{sec:appendix:queueing}

This appendix supplies the latency accounting and reclamation example motivating Section~\ref{sec:motivation_observations}, then specifies the joint selector's objective for a fixed visible horizon.

\subsection{Sample-Path Latency Accounting}
\label{sec:appendix:latency_identity}

Fix a nonempty horizon $\mathcal H_t$ of $n$ requests and an execution following candidate $a$ and its specified continuation. Let $T_{i,t}^{\mathrm{rem}}(a)$ be request $i$'s remaining time to its first returned token. For elapsed time $x\geq0$, define
\begin{equation}
\label{eq:unfinished_population}
U_t^a(x)=\sum_{q_i\in\mathcal H_t}
\mathbf 1\{T_{i,t}^{\mathrm{rem}}(a)>x\}.
\end{equation}
This population includes admitted requests still awaiting their first token. Each indicator integrates to its request's remaining TTFT, giving
\begin{equation}
\label{eq:horizon_area_identity}
\overline T_t^{\mathrm{rem}}(a)
=\frac{1}{n}\sum_{q_i\in\mathcal H_t}T_{i,t}^{\mathrm{rem}}(a)
=\frac{1}{n}\int_0^\infty U_t^a(x)\,dx.
\end{equation}
With finite conditional mean, taking expectations yields
\begin{equation}
\label{eq:expected_unfinished_area}
J_t(a)=\frac{1}{n}\int_0^\infty
\mathbb E[U_t^a(x)\mid s_t]\,dx,
\end{equation}
where $J_t(a)$ is the objective in Eq.~\eqref{eq:horizon_objective}. Interchanging expectation and integration is valid because the integrand is nonnegative. The identity holds for continuous batching and overlapping execution without a stationary-arrival assumption; it does not make service-time predictions exact.

\begin{figure*}[!t]
\centering
\includegraphics[width=0.85\linewidth]{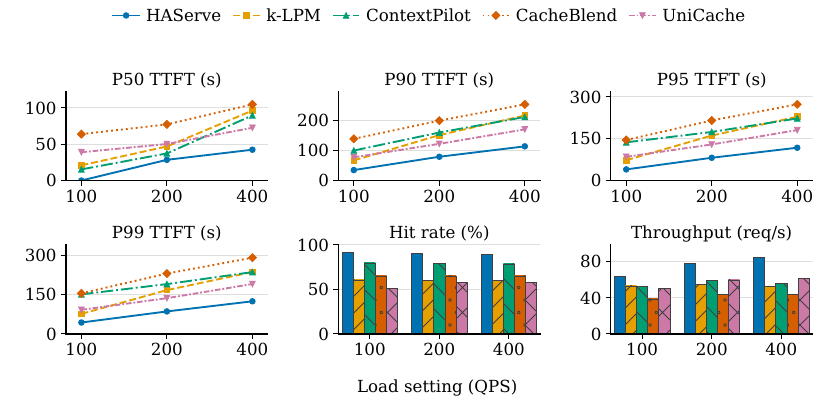}
\caption{\textbf{HAServe retains serving gains with two TP=4 replicas.} Six panels compare four TTFT percentiles, exact-prefix cache hit rate, and throughput on the mixed workload with Qwen3-4B-Instruct-2507 on eight H100 GPUs in total. Load settings denote the combined reference rate across both replicas. All values are means over five random seeds.}
\Description{Six panels compare HAServe, k-LPM, ContextPilot, CacheBlend, and UniCache at load settings of 100, 200, and 400 QPS, using line charts for P50, P90, P95, and P99 TTFT and grouped bars for cache hit rate and throughput.}
\label{fig:eval:two_replica}
\end{figure*}

\subsection{Queue-Weighted Wave Cost}

Partition $\mathcal H_t$ into prefill waves $\mathcal B_{t,w}^a$ under a feasible candidate $a$. Missing-token work in wave $w$ is
\begin{equation}
\label{eq:visible_work}
D_{t,w}^a=\sum_{q_i\in\mathcal B_{t,w}^a}
\bigl(\ell(z_i^a)-\widehat L_{i,t}^a\bigr),
\end{equation}
where $\ell(\cdot)$ counts tokens and $\widehat L_{i,t}^a$ predicts the reusable-prefix length at service after preceding construction and reclamation. In the serial-wave abstraction, a request in wave $k$ completes prefill after shared initial blocking $c_t(a)$ and all wave durations through $k$. Its remaining TTFT is
\begin{equation}
\label{eq:request_wave_cost}
T_{i,t}^{\mathrm{rem}}(a)=c_t(a)+\sum_{w\leq k}S_{t,w}^a+T_{i,t}^{\mathrm{1tok}}(a),
\end{equation}
where $T_{i,t}^{\mathrm{1tok}}(a)$ is the first-token interval from Eq.~\eqref{eq:ttft_decomp}. Let $N_{t,w}^a$ count requests in wave $w$ or later. Summing over requests counts $S_{t,w}^a$ exactly $N_{t,w}^a$ times, giving
\begin{equation}
\label{eq:horizon_wave_delay}
J_t^{\mathrm{wave}}(a)=c_t(a)
+\frac{1}{n}\sum_wN_{t,w}^aS_{t,w}^a
+\overline T_t^{\mathrm{1tok}}(a),
\end{equation}
Here $\overline T_t^{\mathrm{1tok}}(a)$ is the mean first-token interval. Shortening a wave by $\delta$ reduces mean remaining TTFT by $N_{t,w}^a\delta/n$ when other terms are fixed.

This identity assumes serial, nonoverlapping waves. Its online use requires predictions of missing work and service time. In the approximation $S_{t,w}^a\approx D_{t,w}^a/C_{t,w}^a+\eta_{t,w}^a$, $D_{t,w}^a$ is missing-token work, $C_{t,w}^a$ is the effective prefill rate, and $\eta_{t,w}^a$ captures additional wave overhead. Effective rates depend on context lengths, batch composition, chunk limits, and active decode. Continuous batching and prefill--decode interference require backend calibration~\citep{yu22orca,agrawal24sarathi,zhong24distserve}.

Matches follow the evolving cache after construction and reclamation. Rewriting a legal prompt requires matching the new token path, and concurrent requests cannot share cold-prefix computation without backend support. New arrivals are outside the compared horizon and may perturb realized service. Already accrued waiting is common to candidates, while blocking control time enters $c_t(a)$.

\subsection{Reclamation-Induced Ranking Reversal}
\label{sec:appendix:ranking_reversal}

\noindent\textbf{Proposition.} A ranking that preserves initial cache matches can reverse the order of feasible joint admission--retention choices, even for two visible requests.

\emph{Construction.} Consider requests $A,B$ released at the same decision time and a capacity of two KV blocks. Initially, one block holds $A$'s reusable prefix $p_A$, with no active reference. With $p_A$, $A$ needs one additional block and takes time $u>0$. Without it, $A$ occupies two blocks and takes $u+\delta$, where $\delta>0$. Request $B$ occupies two blocks, takes time $v>0$, and shares no reusable prefix with $A$. Thus admitting $B$ first must evict $p_A$. Service times are state-dependent given quantities, not assumed proportional to KV footprints.

Execution is serial and nonpreemptive, with no other active requests or new arrivals. Each request ends at its first returned token, making its KV reclaimable for the next request. There is no continued decode, offloading, prefetching, or external restoration of $p_A$ before $A$ executes. A cache miss still permits normal recomputation. Control and other delays are zero, and neither order is excluded by the common age guard. Retaining $p_A$ means keeping it until $A$ consumes it, after which its space may be reclaimed for $B$. The sole remaining request follows the common FIFO/LRU continuation.

\emph{Proof.} The three feasible initial choices have mean remaining TTFT
\begin{equation}
\label{eq:ranking_reversal_costs}
\begin{aligned}
\overline T_{A\to B}^{\mathrm{keep}}&=(2u+v)/2,\\
\overline T_{A\to B}^{\mathrm{evict}}&=(2u+2\delta+v)/2,\\
\overline T_{B\to A}^{\mathrm{evict}}&=(2v+u+\delta)/2.
\end{aligned}
\end{equation}
Admitting $B$ while retaining $p_A$ requires three blocks and is infeasible. If $0<u-v<\delta$, evicting before serving $A$ adds $\delta$ to the mean, while serving $B$ first adds $(v-u+\delta)/2>0$. Hence serving $A$ first with its prefix retained minimizes the mean among the feasible choices. Yet preserving the initial match assigns $(2v+u)/2$ to $B\to A$ and prefers it because $v<u$. For illustrative values $u=3$, $v=2$, and $\delta=2$, the feasible means are $4$, $6$, and $4.5$, whereas the initial-match estimate for $B\to A$ is $3.5$.

This analytical instance establishes why candidate costs must use post-reclamation matches. It does not assert global scheduling optimality or a tail-latency guarantee for HAServe.

\subsection{Candidate Objective}
\label{sec:appendix:decision_cost}

Let $\mathcal U(s_t)$ be the feasible actions under runtime state $s_t$, including request order, legal serialization, and retention. Feasibility enforces backend capacity and reference constraints and the common age guard. For an action $a\in\mathcal U(s_t)$, the conditional objective is
\begin{equation}
\label{eq:horizon_objective}
J_t(a)=\mathbb E\!\left[
\frac{1}{|\mathcal H_t|}\sum_{q_i\in\mathcal H_t}T_{i,t}^{\mathrm{rem}}(a)
\,\middle|\,s_t\right],\qquad |\mathcal H_t|>0.
\end{equation}
The expectation represents uncertain service and interference. The controller uses current observations and completed events, without observing future arrivals. All candidates use the same FIFO/LRU continuation after the current action. For an empty horizon, the backend follows its fallback rule.

The predicted cost in Eq.~\eqref{eq:haserve_admission_cost} counts shared blocking, queue-weighted service, and first-token intervals. Control and serialization costs enter the corresponding intervals; retention-induced recomputation enters once through missing work. Already accrued waiting is common to candidates.

A candidate must improve on the next-feasible original-path FIFO/LRU reference with the same age guard by the calibrated margin $\tau_t$. This margin accounts for empirical prediction uncertainty but provides no unconditional improvement guarantee.

\section{Effect of Prefix Organization on Answer F1}
\label{sec:appendix:organization_quality}

Exact-path grouping leaves prompts unchanged. Permitted alignment preserves segment contents, system instructions, and fixed history, but changes evidence positions and can affect answer quality~\citep{liu24lostmiddle,contextpilot2026}.

The Qwen3-4B-Instruct-2507 F1 comparisons in Section~\ref{sec:eval:quality_overhead} measure full-system effects on MultiHopRAG and LooGLE. They do not isolate alignment or establish Llama2-13B quality. HAServe restricts alignment to application-permitted orders and uses exact-path grouping without alignment when segment annotations are absent.
\section{Multi-Replica Serving}
\label{sec:appendix:multi_replica}

\noindent\textbf{Setup.} We evaluate Qwen3-4B-Instruct-2507 in BF16 on two independent replicas, each using tensor parallelism across four H100 GPUs (TP=4), for eight GPUs in total. All five methods use the same round-robin request assignment and resource budgets. Each replica has 80 request slots and an 8{,}192-token prefill budget, and maintains its own cache across the mixed workload's stages. The load labels of 100, 200, and 400 QPS denote the combined reference rate across both replicas for the initial RAG stage, with all stage clocks scaled uniformly.

\noindent\textbf{Measurement.} For each run, we pool requests from both replicas before computing TTFT percentiles, weight cache hit rates by input-token counts, and measure completed-request throughput over a common interval that includes draining the request cohort. Every reported value is the arithmetic mean of five measurements using different random seeds.

\noindent\textbf{Results.} Figure~\ref{fig:eval:two_replica} reports all three load settings. Relative to the strongest baseline for each metric at the same load, HAServe reduces P99 TTFT by 34.6--43.1\%, increases throughput by 19.9--37.4\%, and raises cache hit rate by 11.26--11.77 percentage points. The throughput gain grows from 19.9\% at 100 QPS to 37.4\% at 400 QPS, while HAServe sustains an 89.71--91.35\% cache hit rate across the three settings. These results show that HAServe retains its serving benefits when scheduling and cache management operate independently within two replicas under a shared routing policy.

\end{document}